%% file: MAIN.tex
\documentclass{ieeeaccess}
\usepackage{cite}
\usepackage{amsmath,amssymb,amsfonts}
\usepackage{textcomp}
\usepackage{algorithm}
\usepackage{algpseudocode}
\usepackage{booktabs}
\usepackage{subfig}
\usepackage{graphicx}
\usepackage{balance}
\usepackage{bbm}
\usepackage{hyperref}
\hypersetup{colorlinks=true, linkcolor=black}
\usepackage{soul}
\usepackage{comment}
\usepackage[table]{xcolor} 
\newcommand\hlt[1]{%
  \bgroup
  \hskip0pt\color{red!80!black}%
  #1%
  \egroup
}

\usepackage{upgreek}

\newcommand{\fbseries}{\unskip\setBold\aftergroup\unsetBold\aftergroup\ignorespaces}

\newcommand{\setBoldness}[1]{\def\fake@bold{#1}}

\def\BibTeX{{\rm B\kern-.05em{\sc i\kern-.025em b}\kern-.08em
    T\kern-.1667em\lower.7ex\hbox{E}\kern-.125emX}}

\newcommand{\SystemName}{KappaFace}

\newcommand\Tstrut{\rule{0pt}{2.6ex}}       
\newcommand\Bstrut{\rule[-1.1ex]{0pt}{0pt}} 

\begin{document}
\history{Date of publication xxxx 00, 0000, date of current version xxxx 00, 0000.}
\doi{10.1109/ACCESS.2023.3338648}

\title{KappaFace: Adaptive Additive Angular Margin Loss for Deep Face Recognition}
\author{\uppercase{Chingis Oinar}\authorrefmark{1} \authorrefmark{*},
\uppercase{Binh M. Le\authorrefmark{2} \authorrefmark{*}, and Simon S. Woo}.\authorrefmark{2}}
\address[1]{Mercari, Tokyo 106-6118, Japan}
\address[2]{
Department of Computer Science and Engineering, College of Computing and Informatics, Sungkyunkwan University, Suwon 16419, South Korea}
\address[*]{Equally contributed.}

\tfootnote{This work was supported in part by the Institute for Information and Communication Technology Planning and Evaluation (IITP) Grant
funded by the Korean Government (MSIT), Graduate School of Convergence Security at Sungkyunkwan University, under Grant 2022-0-01199; in part by the Self-Directed Multi-Modal Intelligence for Solving Unknown, Open Domain Problems, under Grant 2022-0-01045; in part by the AI Platform to Fully Adapt and Reflect Privacy-Policy Changes under Grant 2022-0-00688; in part by the Artificial Intelligence Innovation Hub under Grant 2021-0-02068, in part by the AI Graduate School Support Program at Sungkyunkwan University under Grant 2019-0-00421; and in part by the Advanced and Proactive AI Platform Research and Development Against Malicious Deepfakes under Grant RS-2023-00230337. Also, this work was supported by Korea Internet \& Security Agency (KISA) grant funded by the Korea government (PIPC) (No.RS-2023-00231200, Development of personal video information privacy protection technology capable of AI learning in an autonomous driving environment).}
\markboth
{Chingis \headeretal: KappaFace: Adaptive Additive Angular Margin Loss for Deep Face Recognition}
{Chingis \headeretal: KappaFace: Adaptive Additive Angular Margin Loss for Deep Face Recognition}

\corresp{Corresponding author: Simon S. Woo (e-mail: swoo@g.skku.edu).}

\input{Sections/0_abstract}


\titlepgskip=-15pt

\maketitle

\input{Sections/1_introduction.tex}

\input{Sections/2_related_works.tex}

\input{Sections/3_method.tex}

\input{Sections/4_experiment}

\input{Sections/5_conclusion.tex}

\bibliographystyle{IEEEtran}
\bibliography{IEEEabrv,bibliography}
\begin{IEEEbiography}[{\includegraphics[width=1in,height=1.25in,clip,keepaspectratio]{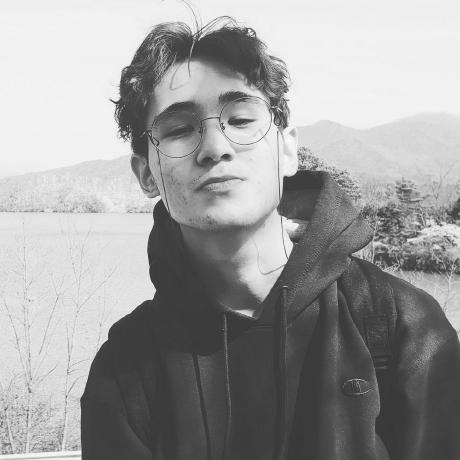}}]{Chingis Oinar} is currently a Machine Learning Engineer at Mercari, Inc., Japan. He obtained the B.S. degree at the College of Computing and Informatics, Sungkyunkwan University. His main research interests include pattern recognition, machine learning, representation learning.
\end{IEEEbiography}

\begin{IEEEbiography}[{\includegraphics[width=1in,height=1.25in,clip,keepaspectratio]{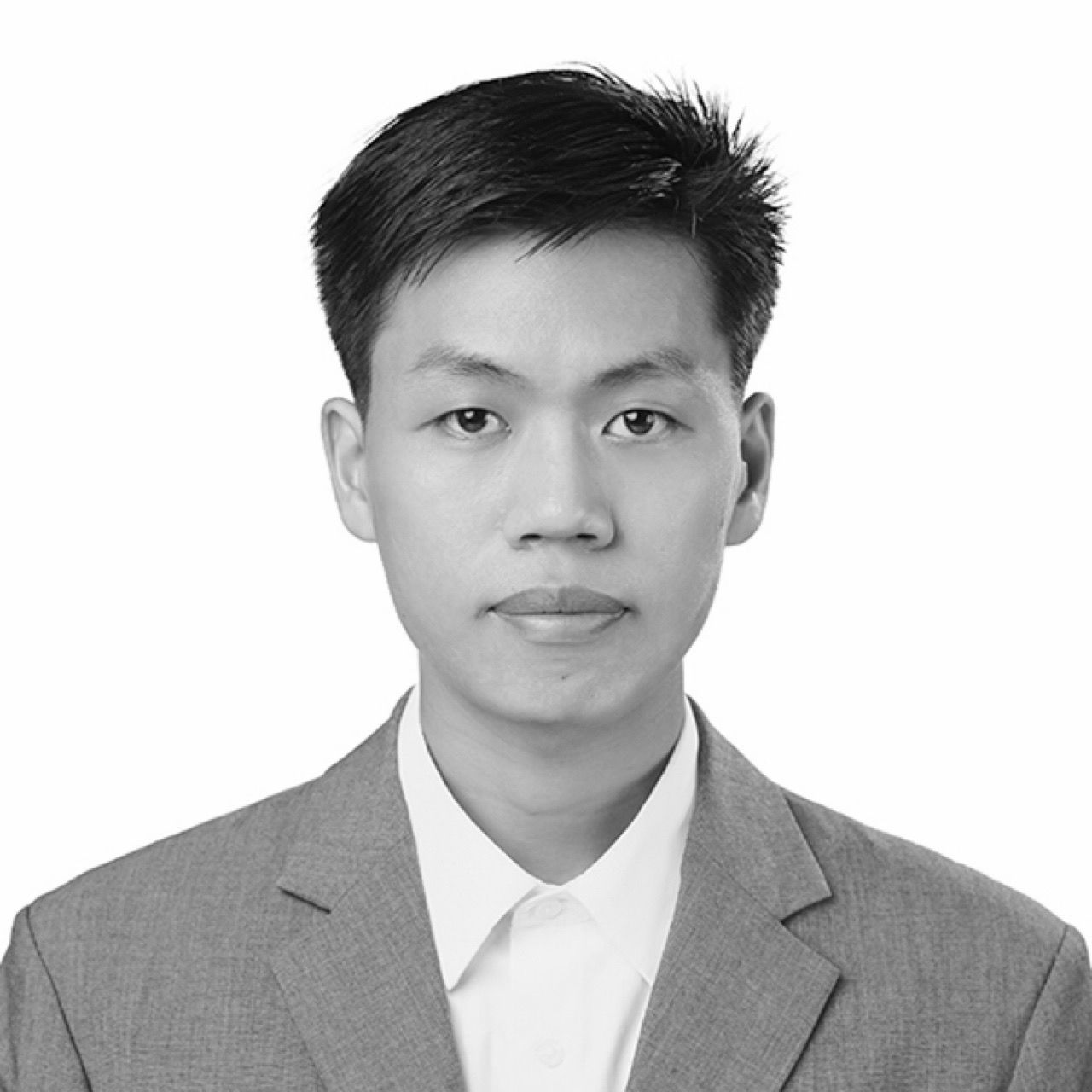}}]{Binh M. Le} received the B.S. degree from the University Sciences, Vietnam National University. He is currently pursuing Ph.D. degree at the College of Computing and Informatics, Sungkyunkwan University. His main research interests include pattern recognition, deepfake detection, and adversarial training. More details about his research and background can be found at \url{https://sites.google.com/view/binhminhle/home}.\end{IEEEbiography}

\begin{IEEEbiography}[{\includegraphics[width=1in,height=1.2in,clip,keepaspectratio]{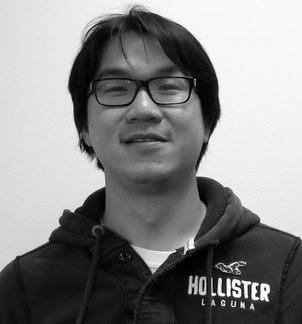}}]{Simon S. Woo} received his Ph.D. in Computer Science from the University of Southern California, Los Angeles/Information Sciences Institute (USC/ISI). He earned his B.S. in Electrical Engineering degree (B.S.EE) from Univ. of Washington (UW), Seattle, and M.S. in Electrical and Computer Engineering degree (M.S.ECE) from Univ. of California, San Diego (UCSD). Currently,  he is an Associate Professor with the College of Computing and Informatics, Sungkyunkwan University (SKKU), Suwon, South Korea. He has published a number of papers in peer-reviewed journals and conference proceedings, such as NeurIPs, AAAI, IJCAI, ACM MM, WWW, CIKM, etc. His research interests include Multimedia Forensics including deepfake detection, AI security, and anomaly detection. Before joining SKKU, he was a Member of Technical Staff for 9 years at the NASA’s Jet Propulsion Laboratory (JPL), Pasadena, CA, conducting research in satellite communications, computer networking, AI, and security.  Also, he is the organizer for the 2022-2024 Workshop on the Security Implications of Deepfakes and Cheapfakes (WDC) at AsiaCCS. More information can be found at \url{https://dash-lab.github.io/About/}.\end{IEEEbiography}

\EOD

\end{document}

%% file: Sections/0_abstract.tex
\begin{abstract}
Feature learning is a widely used method for large-scale face recognition tasks. Recently, large-margin softmax loss methods have demonstrated significant improvements in deep face recognition. However, these methods typically propose fixed positive margins to enforce intra-class compactness and inter-class diversity, without considering imbalanced learning issues that arise due to different learning difficulties or the number of training samples available in each class.  {This overlook not only compromises the efficiency of the learning process but, more critically, the generalization capability of the resultant models.} 
 {To address this problem,}  we introduce a novel adaptive strategy called KappaFace, which modulates the relative importance of each class based on its learning difficulty and imbalance. {Drawing inspiration from the von Mises-Fisher distribution, KappaFace increases the margin values for the challenging or underrepresented classes and decreases that of more well-represented classes.}
{Comprehensive experiments across eight cutting-edge baselines and nine well-established facial benchmark datasets strongly confirm the advantage of our method. Notably, we observed an enhancement of up to 0.5\% on the verification task when evaluated on the IJB-B/C datasets. In conclusion, KappaFace offers a novel solution that effectively tackles imbalanced learning in deep face recognition tasks and establishes a new baseline.}
\end{abstract}

%% file: Sections/1_introduction.tex
\section{Introduction}
Deep Convolutional Neural Networks (DCNNs) have showcased remarkable advancements in the realm of image representation learning, making them extensively adopted for a set of image-related tasks, including face recognition and verification. Recently, it has also been observed that the traditional softmax loss fails to produce highly discriminative feature vectors ~\cite{wang2017normface}. Thus, Deep Metric Learning (DML) has received massive attention and has been used for a variety of tasks, especially face recognition~\cite{schroff2015facenet, wen2016discriminative}. However, due to the high computational costs of the previously proposed methods, such as Triplet Loss, on larger datasets, several variants have been proposed to improve the discriminative power of the softmax loss ~\cite{deng2019arcface, wang2018cosface, liu2017sphereface, huang2020curricularface, wang2017normface, liu2019fair}. Therefore, current state-of-the-art methods focus on modulating positive margins to enforce intra-class compactness and inter-class diversity, such as ArcFace ~\cite{deng2019arcface}, CosFace ~\cite{wang2018cosface}, and SphereFace ~\cite{liu2017sphereface}. Moreover, current trends in deep face recognition involve performing face identification during training and face verification during the inference stage, making the softmax-based loss a favorable setting. In the context of deep face recognition, the softmax-based loss function is used to train a model to recognize the specific identity of a person in an image. The model is trained on a dataset of labeled images, where each image is labeled with the identity of the person in the image. The model learns to associate certain features in the image with certain identities. Furthermore, the fixed margins significantly improved the generalization ability of DCNN, achieving great performance on a series of major face benchmarks, namely LFW ~\cite{huang2008labeled}, CFP-FP ~\cite{sengupta2016frontal}, CPLFW ~\cite{zheng2018cross}, AgeDB ~\cite{moschoglou2017agedb}, CALFW ~\cite{zheng2017cross}, IJB-B ~\cite{whitelam2017iarpa} and IJB-C ~\cite{maze2018iarpa}. Moreover, margin-based losses have been also introduced in other domains of deep face recognition tasks, such as kinship verification, where kinship verification involves predicting whether two individuals are related, which might include parent-child or sibling relationships. Duan et al. propose a method to tackle the issue of distribution difference between related identities by enforcing larger distances between negative pairs through a constant margin~\cite{zhang2020advkin}.
\begin{figure}[t!]
\centering
\includegraphics[width=8.4cm]{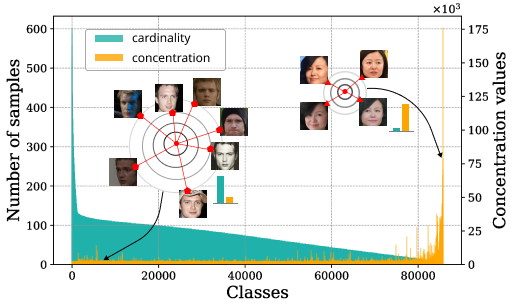}
\caption{Distribution of cardinalities of classes in MS1MV2 dataset and their concentration values in the embedding space produced by ArcFace. The concentration values indicate how close the embeddings are scattered around their corresponding centroids. The produced concentration values are affected by both the number of samples available and the variation in terms of image quality, facial angles and ages, making some identities difficult to learn ({left and right sets}). Our proposed~\SystemName~considers these two contribution factors and adjusts the margin for each class in order to train a robust facial embedding model.} 
\label{fig:arcface_distribution}
\vspace{-10pt}
\end{figure}

\begin{figure*}[t!]
\centering
\includegraphics[width=6in]{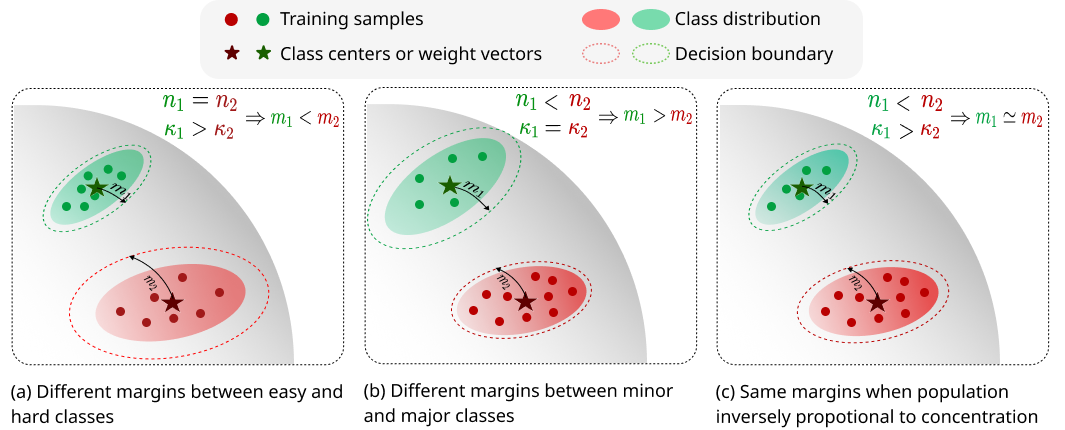}
\caption{\textbf{Effects of our adaptive strategy in KappaFace.} {Here, $n_1$, $k_1$ and $n_2$, $k_2$ are the total number of samples (population) and class concentrations for some classes labeled as $1$ and $2$, also colored as green and red, respectively.} (a) We prioritize difficult classes by intensifying margins for low-concentration classes {(in red)}. (b) Minority classes {(in green)} receive relatively larger margins than majority classes do {(in red)}. (c) Finally, we assign similar importance provided the population of one class is proportional to the concentration of another.}
\label{fig:margin_change}
\end{figure*}

{The margin value determines how far apart the feature embeddings of different classes should be from each other. A larger margin value will result in a larger angular distance between the feature embeddings of different classes, and 
 vice versa}~\cite{deng2019arcface}. {Nevertheless, a larger margin value may lead to better performance on the training set, but may also lead to overfitting and poorer generalization to unseen data. On the other hand, a smaller margin makes the model too lenient in its decision boundary and may not be able to effectively distinguish between positive and negative samples}~\cite{deng2019arcface, wang2018additive}. {Recent works, such as CurricularFace }\cite{huang2020curricularface}, {demonstrated previous large margin methods being inefficient for datasets having issues such as class imbalance or high learning difficulty, due to samples that vary in different aspects, as shown in Fig.} \ref{fig:arcface_distribution}. {In fact, the class imbalance turns out to be a major challenge for deep face recognition models. Thus, the fixed margin methods were shown to produce biased decision boundaries, as demonstrated by Liu et al. }\cite{liu2019fair}{, and minority classes poorly describe the true feature space.}


{Despite the saturation of performance metrics on deep face datasets and the intricate state-of-the-art advancements, our work seeks to break through these barriers. While the challenges of imbalanced learning is well-acknowledged, our novel contribution in this paper is the introduction of the Adaptive Additive Angular Margin Loss or KappaFace, which tackles imbalance in both class population and class learning difficulty. Through a clear conceptual differentiation depicted in Fig. {\ref{fig:margin_change}}, we tackle the issues associated with fixed margins by breaking down the problem into two distinct sub-cases: class difficulty (Fig. {\ref{fig:margin_change}}-a) and class size (Fig. {\ref{fig:margin_change}}-b).}  In this paper, we propose our novel method, Adaptive Additive Angular Margin Loss or KappaFace, to further tackle these challenges. 
As illustrated in  Fig. \ref{fig:margin_change}{, we resolve the challenges  {associated with fixed margins} by decomposing the problem into \textit{two} sub-cases based on the difficulty }(Fig. \ref{fig:margin_change}-a){ and cardinality} (Fig. \ref{fig:margin_change}-b){ of a class. In particular, during the training, we emphasize hard-to-learn classes by increasing margins for classes with high dispercity (or low concentration).  {Likewise}, for underrepresented (or small) classes, we increase their margins to compensate for data scarcity.} {Our end goal is to encourage well-learned identities with closely projected samples, thus achieving invariant representations. We demonstrate, based on our novel observations, that identities with samples showing minimal facial angles and age variations are more likely to result in invariant representations and are inherently easier to learn.}

 {In contrast to recent approaches, our novel method, denoted as \mbox{\SystemName}, uniquely employs the von Mises-Fisher distribution (vMF) to create a method that leverages both the concentration and population parameters of a training dataset. This design results in an adaptive class-based margin loss.} Especially, the vMF is utilized to model the spherical distribution of instances of given identities, as well as to estimate the class concentration parameter, which reflects the difficulty in learning due to variations in facial angles, ages, and resolutions. {Building upon the family of margin-based objectives, first introduced by by ArcFace ~\mbox{\cite{deng2019arcface}}, our core objective is to refine and balance these margins across classes. This approach strategically enhances the generalization capabilities, particularly for datasets with class imbalance.}
As shown in Fig. \ref{fig:margin_change} our approach behaves differently across multiple scenarios, which include classes of different learning difficulties as well as varying populations, making our approach flexible and effective for highly imbalanced training datasets. Our main contributions are summarized as follows\footnote{Our code is available here: \url{https://github.com/chingisooinar/KappaFace}}:

\begin{itemize}
    
    \item We propose a novel method, KappaFace, which adjusts the class importance by taking both its learning difficulty and the imbalance degree into account. {Leveraging the capabilities of the vMF distribution},  KappaFace relaxes the margins of high-concentration classes and intensifies that of poorly learned and underrepresented classes. 

    \item For modeling hyperspherical distributions, 
    we propose updating margins through concentration and population parameters during training, which does not produce any extra computational costs in the inference stage, using the momentum encoder. We also provide a comparison with the memory buffer based approach, which might be more desirable with smaller datasets.   
    
    \item We evaluate our method on a series of popular face recognition benchmarks and demonstrate that our KappaFace consistently outperforms other state-of-the-art counterparts.  
\end{itemize}

 {The structure of this paper is described as follows. In Section {\ref{sec:related_work}}, we discuss closely related works and elucidate our sources of inspiration. In Section {\ref{sec:method}}, we provide a detailed description of our proposed approach to address the challenge of imbalanced training data by leveraging the vMF distribution through two modeling strategies: the memory buffer and the momentum encoder. In Section {\ref{sec:exp_results}}, we assess our method, KappaFace, comparing it with the latest state-of-the-art methods across nine facial benchmarks and provide additional ablation studies. We conclude our research in Section {\ref{sec:conclusion}} and suggest avenues for future research.}

%% file: Sections/2_related_works.tex
\section{Related works}
\label{sec:related_work}
{
In this section, we provide a concise review of research relevant to our study, focusing on the following domains: deep metric learning, normalization, spherical distribution, margin-based loss functions, and issues of class imbalance.}

\textbf{Deep Metric Learning.} Deep Metric Learning (DML) is currently widely adopted for face recognition and verification tasks as it is observed to produce highly efficient feature representations compared to the traditional softmax loss ~\cite{hu2014discriminative, liu2017adaptive}. {As DML has become a de facto approach for face recognition and verification, other related tasks, where earlier approaches relied on handcrafted features~\mbox{\cite{turk1991eigenfaces,belhumeur1997eigenfaces,ahonen2004face,saeed2024framework, kareem2024face}}, have also been receiving attention and benefit from the robust, discriminative features learned by novel algorithms~\mbox{\cite{ngo2020facial}}}. Thus, with the increasing attention towards DML, various methods have been proposed in recent years, albeit some of the eminent works in {this} area include FaceNet~\cite{schroff2015facenet}, Center Loss ~\cite{wen2016discriminative}, Circle Loss ~\cite{sun2020circle}, L-Softmax Loss ~\cite{liu2016large}, N-pair Loss ~\cite{sohn2016improved}, AdaReg~\cite{li2021adaptively}, CRS-CONT~\cite{li2022crs}, and Ada-CM~\cite{li2022towards}. 
However, the above methods, such as Triplet Loss, N-pair Loss, and Circle Loss, which require mining of both negative and positive samples, become computationally burdensome as the size of the dataset increases.


 \textbf{Normalization and Spherical Distribution.} Normalization plays a crucial role in current state-of-the-art deep face recognition methods. The majority of works apply $l_2$ normalization on the projection matrix and the feature vectors to embed the latter onto the unit hypersphere~\cite{deng2019arcface, liu2017sphereface, wang2018cosface, liu2019fair, huang2020curricularface}. In this fashion, the inner product is replaced by cosine similarity within the softmax loss. Interestingly, the $l_2$ normalization was observed to significantly contribute to the model performance as shown by NormFace ~\cite{wang2017normface}. 
 
 Furthermore, it was reported that normalization reduces prior caused by the training data imbalance issue ~\cite{liu2017sphereface}. Therefore, modeling the hyperspherical distribution of facial features, particularly the von Mises-Fisher distribution, has drawn attention from recent works~\cite{hasnat2017mises, zhe2019directional, li2021spherical}. Specifically, they produce the belongingness probability from a predicted or mini-batch estimated concentration. Although vMF Loss~\cite{zhe2019directional} mentions the class concentration parameter, it simply uses a fixed value across classes as a scaling parameter while trying to minimize the distance between the visual representation and the mean representation of the corresponding class.
 In contrast, we propose two alternative approaches, namely, the memory buffer and momentum encoder, to manage the concentrations of all classes simultaneously. Thereby, we can modulate the concentrations' diversity and, generally, adjust our loss upon the relative learning difficulty and imbalance of an individual class.

\textbf{Margin-based Loss Functions.} Given that the features learned using the traditional softmax loss have not been discriminative enough for practical face recognition tasks, state-of-the-art (SOTA) deep face recognition methods often incorporate margin-based loss functions, as demonstrated in~\cite{deng2019arcface, liu2017sphereface, wang2018cosface, liu2019fair, huang2020curricularface}. {Furthermore, margin-based loss functions have also been introduced in semi-supervised learning. Ada-CM introduces adaptive margins to effectively leverage unlabeled  {data} by learning facial expression features}~\cite{li2022towards}. 

However, there have been several challenges reported with margin-based losses, {such as} dataset imbalance or sample learning {difficulty}. Based on this observation, there are several methods that aim to overcome these challenges, such as Fair Loss ~\cite{liu2019fair} or CurricularFace ~\cite{huang2020curricularface} by introducing Reinforcement Learning (RL) agent and negative margin, respectively. However, the introduction of an additional network to train on top of a backbone is not only a non-trivial task, but also a computationally intense task for training. Meanwhile, though the negative margin introduced by CurricularFace addresses the challenge of sample learning difficulty, it does not take the class imbalance into account. Thereby, our proposed KappaFace aims to address both challenges, class learning difficulty and its imbalance, with no additional trainable parameters.

\textbf{Class Imbalance.} Class imbalance is a common problem in deep face recognition. Traditional approaches to tackle this issue include data re-sampling ~\cite{he2009learning} and cost-sensitive learning ~\cite{krawczyk2014cost, tang2008svms}. Furthermore, there are also methods, such as Class-Balanced Loss ~\cite{cui2019class}, which introduce modifications directly into the classical softmax loss. Although Class-Balanced Loss might be effective for the classification task in general, it, unfortunately, fails to produce highly discriminative feature vectors, which is vital for the deep face recognition tasks. However, there have been only few methods proposed to address this issue. {Cao et al. propose Label-Distribution-Aware Margin Loss (LDAM), inspired by minimizing a margin-based generalization bound }\cite{cao2019learning}. { Particularly, LDAM modulates margins to enforce greater values for
minority classes, i.e., classes that have a small number of samples. Thus, the margins are determined by class populations. Although KappaFace also tackles class imbalance issue, our method also takes class concentrations into account, which reflects their learning difficulties. Liu et al. propose a method, Fair Loss }~\cite{liu2019fair},{ to modulate margins by introducing a RL agent on top of a backbone. Likewise, Fair Loss might learn to enforce or relax margins based on the learning difficulties. On the other hand, our method does not rely on additional trainable networks, which can significantly reduce the training cost and ensure a high generalization ability.}

%% file: Sections/3_method.tex
\section{Proposed method}
\label{sec:method}
{In this section, we first describe the conventional objective functions utilized with margin values. Subsequently, we briefly explain how we model the distribution of each class on a sphere using the vMF distribution and provide details on how our proposed KappaFace efficiently estimates the concentration parameters, $\kappa$, of these distributions. Finally, we outline our approach that leverages the vMF distribution to adjust our margin loss, and we also include a gradient analysis of our loss function.}

\begin{figure}[t!]
\centering
\includegraphics[width=7.4cm]{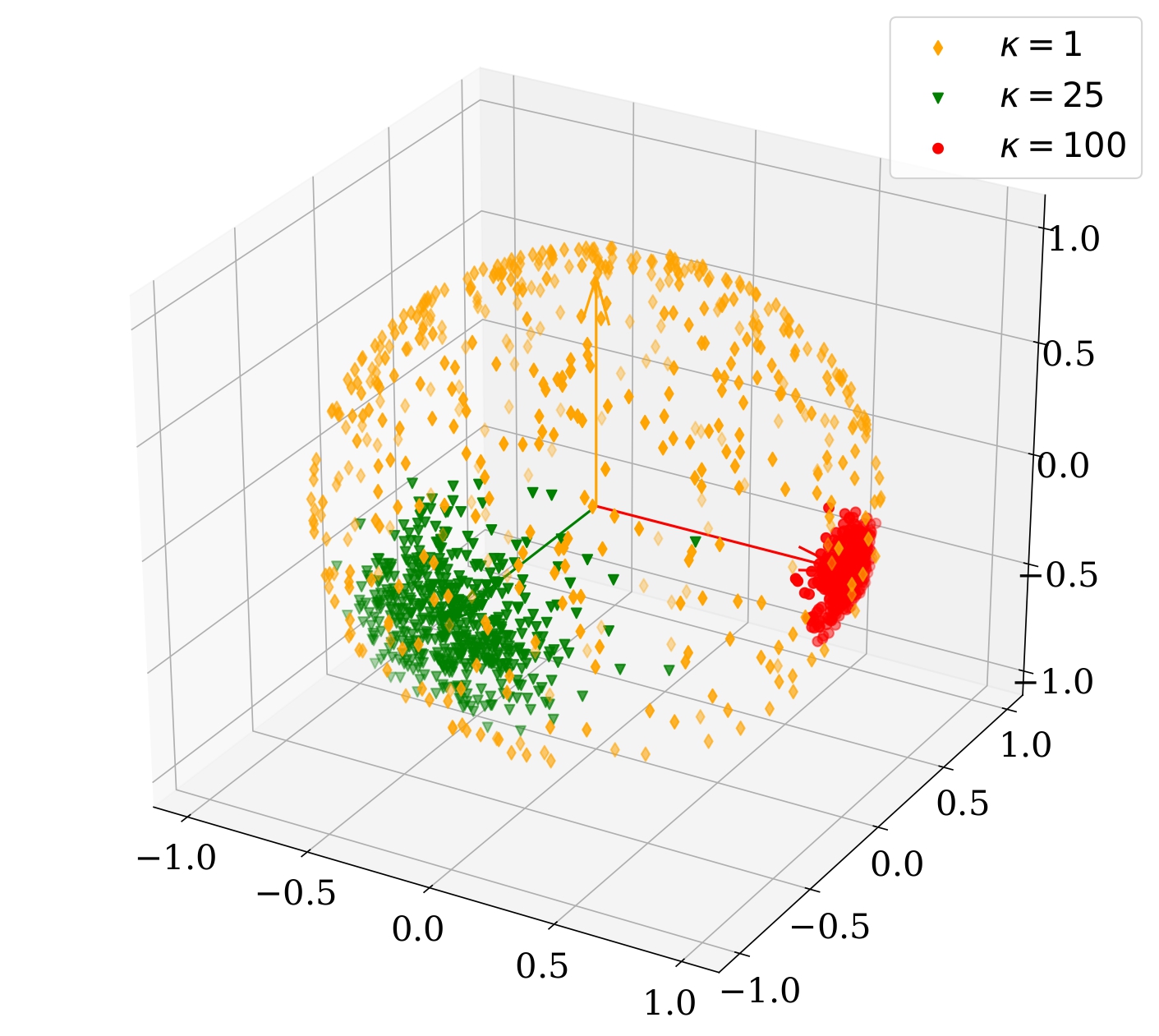}
\caption{Illustration of von Mises-Fisher (vMF) distribution on 3-dimensional unit sphere {with different $\kappa$ concentration parameters.} The yellow-, green-, and red-colored classes have their concentration parameter $\kappa$ increasing, meaning their samples in one class become more compact.} 
\label{fig:vmf}
\end{figure}

\begin{figure*}[!t]
\centering
\includegraphics[width=17cm]{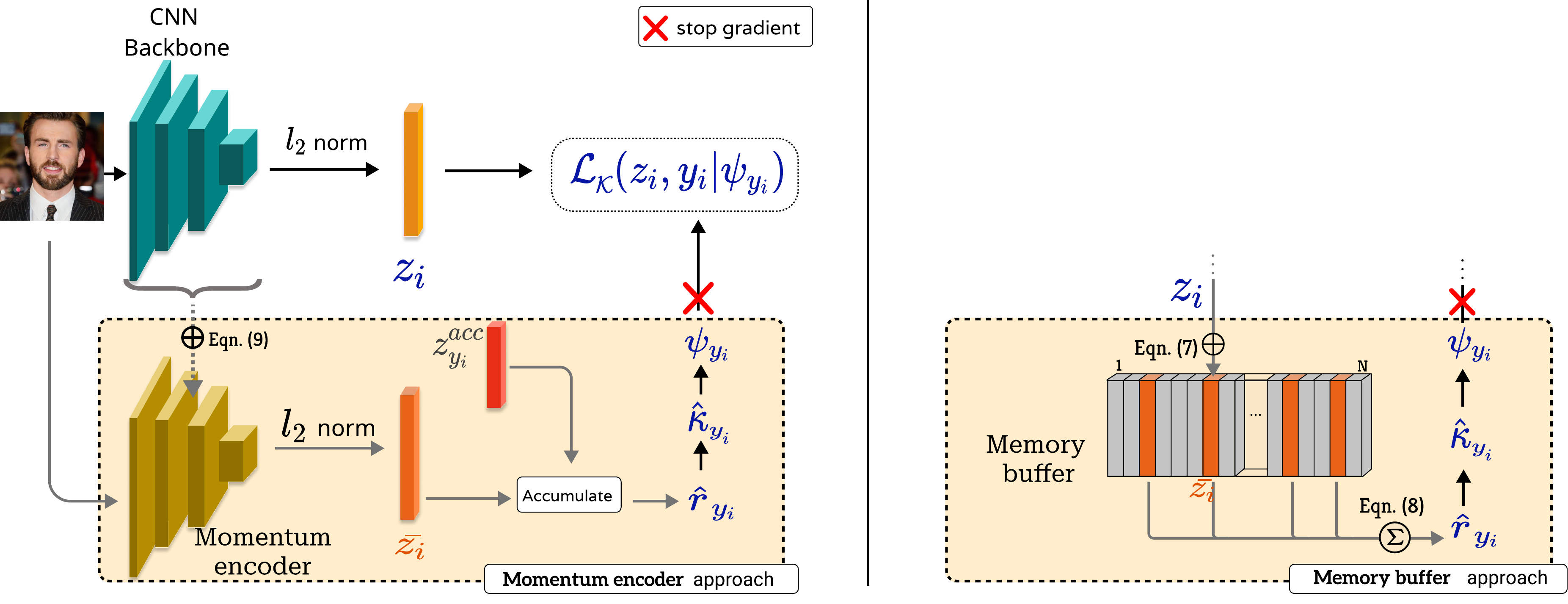}
\caption{Our overall end-to-end diagram of KappaFace with the two proposed concentration, $\hat{\kappa}$, estimation approaches. The \textbf{\texttt{momentum encoder}} (\emph{bottom-left panel}) accumulates the representations for each identity $y_i$, initialized as zero vector $z_{y_i}^{acc}$, to efficiently compute numerator of Eq. \ref{eqn:mean_vmf_modified}. In the \textbf{\texttt{memory buffer}} (\emph{bottom-right panel}), hyper-spherical embedding vector $z_i$ is fed into our memory buffer to update corresponding  $\hat{\kappa}_{y_i}$, producing scaler $\psi_{y_i}$. In the inference stage, the memory buffer and momentum encoder are simply dropped; hence the backbone alone is used to derive feature representations.} 
\label{fig:overall}
\end{figure*}
\subsection{Preliminary Information on Loss Functions}

{We begin by defining the notions used in our paper. Let $\mathbf{z}_{i}\in \mathbb{R}^{d}$ be the embedding of the \textit{i}-th sample corresponding to the class $\mathbf{y}_{i}$, where $\mathbf{y}_{i} \in \{ 1,...,C \}$. Let $\mathbf{W}_{j}\in \mathbb{R}^{d}$ represent the \textit{j}-th column of the weight matrix $\mathbf{W}\in \mathbb{R}^{d \times C}$, and $\mathbf{b}_{j}$ denote the bias term.}

 In contemporary representation learning practices, the bias term $\mathbf{b}_{j}$ is set to 0, whereas the weight matrix $\mathbf{W}$ and the deep feature $z_i$ are normalized using ${l}_{2}$ normalisation method as in ~\mbox{\cite{deng2019arcface, wang2018cosface, liu2017sphereface}}.  However, conventional softmax loss often results in embeddings with low discriminative characteristics. As a remedy, a plethora of margin-based techniques have been introduced. Their general form can be expressed as:
\begin{equation}
\mathcal{L_M} =   -\log \frac{e^{s(P(\cos{\theta_{y_{i}}}))}}{e^{s(P(\cos{\theta_{y_{i}}}))} + \sum_{j=1, j\neq {y}_{i} }^{C}e^{s(\cos{\theta_{j}})}},
\end{equation}
where $\theta_{j}=\angle (\mathbf{W}_{j}, z_i)$, \emph{s} is a scaling factor, $P(\cos{\theta_{y_{i}}})$ is a function to control the positive margin. For instance, ArcFace~\cite{deng2019arcface} introduces the function $P(\cos{\theta_{y_{i}}})$ as $\cos(\theta_{y_{i}} + m)$, where \emph{m} is an additive angular margin, whereas CosFace~\cite{wang2018cosface} and SphereFace~\cite{liu2017sphereface} introduce additive cosine and multiplicative angular
margins, respectively. Thus, utilizing the margin penalties within the softmax loss can effectively enforce intra-class compactness and inter-class diversity by penalizing a target logit.

However, the aforementioned methods introduce a fixed margin for all classes without taking the class imbalance into account, which is a common practical challenge for deep face recognition models\cite{liu2019fair}. Based on this observation, we propose a novel approach that dynamically adjusts the relative importance based on the difficulty of each class and the degree of class imbalance.

\subsection{The von Mises-Fisher distribution}

The von Mises-Fisher distribution (vMF) is a probability distribution on the \emph{d}-dimensional unit sphere $\mathbb{S}^{d-1}$, which has the density function defined as follows~\cite{mardia2009directional}:
\begin{equation}
    p (x | \epsilon, \kappa) = \frac{\kappa ^{d/2-1}}{(2\pi) ^{d/2}I_{d/2-1}(\kappa)} \exp (\kappa \epsilon ^{T} x), x \in \mathbb{S}^{d-1},
\end{equation}
where $\epsilon \in \mathbb{S}^{d-1}$ is the mean direction, $\kappa \in \mathbb{R}^{+}$ is the concentration parameter and $I_v$ denotes the modified Bessel function of the first kind at order $v$  \cite{temme1996special}. {Thus, the vMF distribution is a common distribution for directional statistics because it models data that is distributed around a mean direction.} The distribution of samples is uniform on the sphere when $\kappa=0$, while {they are} more concentrated around the mean direction $\epsilon$ as $\kappa$ increases. An illustration of this behavior is provided in Fig.\ref{fig:vmf}, where the red-colored class with the highest concentration has its samples located much closer to its centroid, while the samples in the yellow-colored class are scattered almost uniformly resulting in the smallest concentration. {In other words, the higher the concentration parameter, the more tightly the projections are clustered around the mean direction. }
Thus, we hypothesize that samples in the classes that are easier to learn are scattered around their centers and produce high concentration values. In contrast, the opposite classes, where the classes are difficult to learn contain noise or samples having a high variance in terms of facial angles, ages, and resolutions, result in lower concentration values.

To approximate the value of the concentration parameter from a set of $n$ observation samples, referred to as its population, we adopt a method proposed by Banerjee et al.~\cite{banerjee2005clustering} that follows the maximum likelihood estimation method, and it is given by:

\begin{minipage}{0.45\linewidth}
\begin{equation}
  \hat{\kappa} = \frac{\bar{r} (d - \bar{r}^{2})}{1-\bar{r}^{2}} \label{eqn:kappa}
\end{equation}
\end{minipage}
\hfill
\begin{minipage}{0.45\linewidth}
\begin{equation}
  \text{, where }\bar{r} = \frac{\left\Vert\sum_{i}^{n}x_{i}\right\Vert_{2}}{n}.
\label{eqn:r_bar}
\end{equation}
\end{minipage}
 Hence, the vMF is used for modeling the sample distributions of the classes in the hyperspherical space. Meanwhile, we can use the concentration value $\hat{\kappa}$, hence the name - \SystemName, to modulate the angular margin for samples belonging to a certain class. {Overall, our fundamental idea revolves around incorporation of vMF distribution, specifically concentration parameter ($\hat{\kappa}$), into margin computation. 



\subsection{Calculation of $\kappa$}
 It is important to emphasize that collecting the representation vectors by the end of each training epoch becomes extremely time-consuming, as the size of the training data increases. Furthermore, considering that the model is updated every mini-batch, direct use of intermediate feature vectors $z_i$ will broadly scatter samples belonging to the same class on the hypersphere, distorting its true concentration. In order to efficiently estimate the mean direction vectors $\epsilon$ and concentrations $\kappa_c$, we introduce two following approaches:

\subsubsection{\textbf{Memory buffer}}In the training phase, a memory buffer that is randomly initialized with the size of $\vert\mathcal{D}\vert \times d$, where$\vert\mathcal{D}\vert $ denotes the training set's cardinality. Thus, we employ the Exponential Moving Average (EMA) \cite{li2019gradient} to update feature vectors within the memory. Specifically, let $\bar{z}^{t}_{i}$ be the average of the feature vector $z_i$ for the \emph{i}-th sample at the \emph{t}-th epoch. Hence, we update the memory for the \emph{t+1}-th epoch as follows:
\begin{equation}
\label{eqn:memory}
    \bar{z}^{t+1}_{i} = \alpha \times \bar{z}^{t}_{i} + (1 - \alpha) \times z_{i},
\end{equation}
where $\alpha$ is the momentum parameter.  
Likewise, the vectors in the memory buffer are normalized using $l_{2}$ normalisation. Thus, Eq. \ref{eqn:r_bar} can be restated for the \emph{c}-th class as follows:
\begin{equation}
\label{eqn:mean_vmf_modified}
    \widehat{r}_{c} = \frac{\left\Vert \sum_{i| y_i=c } \bar{z}_{i} \right\Vert_{2}}{n_c},
\end{equation}
where the sum of vectors $\bar{z}_{i}$ is calculated while training within the memory buffer. This allows us to efficiently compute the concentration values $\widehat{\kappa}_c$ for each class at the end of an epoch using Eq. \ref{eqn:kappa}.
\begin{algorithm}[t!]
\caption{\SystemName~margin loss training with {{\texttt{memory buffer}}} and {{\texttt{momentum encoder}}}, respectively. (We use {orange} and {blue} colors to indicate different steps of \textcolor{orange}{{\texttt{Memory buffer}}} and  \textcolor{blue}{{\texttt{Momentum encoder}}} approaches, respectively.) }
\label{alg:kappaface}
\begin{algorithmic}[1]
\Require Embedding network
parameters $\Theta$, last fully-connected layer parameters $\mathbf{W}$, training dataset $\mathcal{D}$, initial margin value $m_0$, temperature value $T$, learning rate $\alpha$.  \textcolor{orange}{{\texttt{Memory buffer}}}:  memory buffer $\mathcal{M}$.  \textcolor{blue}{{\texttt{Momentum encoder}}}: momentum update parameter $m$, calculation head $\mathcal{Z}^{acc} \in \mathbb{R}^{{N_c}\times 512}$.
\State \textcolor{blue}{$\xi =$ copy($\Theta$) }
\While{not converged}
    \For{$x_{i},\textrm{ } y_{i} \in \mathcal{D}$}
        \State \textit{\# Forward sample and embed to unit sphere}
       \State $z_{i} = \Theta(x_i)$, \textcolor{blue}{$\quad \bar{z}_{i} = \xi(x_i)$}
         \State  $z_{i} = z_{i} \textrm{ }/\textrm{ } \Vert z_{i} \Vert _{2}$, \textcolor{blue}{$\quad \bar{z}_{i} = \bar{z}_{i} \textrm{ }/\textrm{ } \Vert \bar{z}_{i} \Vert _{2}$}
         \State $\mathbf{W} = \mathbf{W} \textrm{ }/\textrm{ } \Vert \mathbf{W} \Vert _{2}$
         \State \textcolor{orange}{\textit{\# Update $\mathcal{M}$ by Eq. \ref{eqn:memory}}}
        \State \textcolor{orange}{ $ \mathcal{M}(z_{i},\textrm{ } y_{i})$ }
        \State \textcolor{blue}{\textit{\# Pass $\bar{z}$ to compute class sums}}
        \State \textcolor{blue}{$z^{acc}_{y_i} \quad  += \quad \bar{z}_{i}$ }
        \State \textit{\# Compute $\mathcal{L_K}$ by Eq. \ref{eqn:kappa_loss} and update model}
        \State $loss = \mathcal{L_K}(z_{i}, \textrm{ }y_{i} \vert \textrm{ } \psi_{y_i})$ 
        \State  $\Theta \textrm{ }\gets \textrm{ } \Theta - \alpha \cdot \nabla _{\Theta} loss$  
        \State  $\mathbf{W} \textrm{ }\gets \textrm{ } \mathbf{W} - \alpha \cdot \nabla _{\mathbf{W}} loss$  
        \State \textcolor{blue}{$\xi \textrm{ }\gets \textrm{ } m\xi + (1 - m)\Theta$}
    \EndFor
    \State \textit{\# Update the scaler $\psi$ by Eq. \ref{eqn:kappa_w}, \ref{eqn:sample_w} and \ref{eqn:final_margin}}
    \State \textcolor{black}{$w^{s}, \textrm{ }w^{k} \textrm{ }\gets \textrm{ } \mathcal{M}.update\_weights(T)$} 
    \State \textcolor{black}{$\psi \textrm{ }\gets \textrm{ } \mathcal{M}.scaling\_factor(w^s, w^k)$ }
\EndWhile
\end{algorithmic}
\end{algorithm}
\subsubsection{\textbf{Momentum encoder}} Alternatively, this approach relies on the momentum encoder introduced by He et al \cite{he2020momentum}. Under this setting, our CNN backbone architecture is commonly considered as the online encoder, and it is used to initialize the momentum encoder \cite{grill2020bootstrap}. Subsequently, a momentum update is introduced in order to address the issue of representations’ inconsistency arising due to the naive copying of the online encoder. The momentum update is provided as follows:

\begin{equation}
\label{eqn:momentum_update}
    \xi = m\xi + (1 - m)\Theta,
\end{equation}
where $\xi$ and $\Theta$ are the parameters of the momentum encoder and online encoder, respectively, and $m \in [0, 1)$ is a momentum coefficient. Finally, the representations from the momentum encoder are used in a similar way as shown in Eq.~\ref{eqn:mean_vmf_modified}.

\textbf{Discussion.} Considering that the intermediate representations $\bar{z}_{i}$ are only used to compute corresponding class sums, the momentum encoder might be a desirable architectural choice as the dataset size increases. He et al.~\cite{he2020momentum} also argue that the representations stored in the memory bank are less consistent since they are updated when they are last seen. On the other hand, the momentum update on the encoder is performed every iteration \cite{he2020momentum}. As a result, even though the samples are encoded
by different encoders (in different mini-batches), the difference among these encoders can be reduced. Although He et al.~\cite{he2020momentum} also demonstrate that the momentum encoder provides a better performance under contrastive learning setting, we report the effects of their explicit use with the vMF distribution in this work.

\subsection{\SystemName}
We present our overall end-to-end architecture of KappaFace in Fig.~\ref{fig:overall}. The right-hand side demonstrates how~\SystemName~works with the memory buffer, whereas the left-hand side shows how we can integrate the momentum encoder. Overall, both approaches are used to approximate $\widehat{r}_{c}$, where the numerator in the Eq. \ref{eqn:mean_vmf_modified} is easily calculated on the fly.

Computing individual class margins involves two factors to consider: their imbalance degrees and learning difficulties. Therefore, we present \textbf{class concentration} and \textbf{population weights}, where concentration values are recalculated once an epoch to update the final additive margin values. Initially, the concentration values $\kappa$ are normalized by subtracting mean and dividing by the standard deviation as follows:
\begin{align}
    \mu_{\kappa} &= \frac{\sum_{c=1}^{C}\widehat{\kappa}_{c}}{C},  \\
    \sigma_{\kappa} &= \sqrt{\frac{\sum_{c=1}^{C}(\widehat{\kappa}_{c} - \mu_{\kappa})^{2}}{C}},  \\
    \Tilde{\kappa_{c}} &= \frac{(\widehat{\kappa}_{c} - \mu_{\kappa})}{\sigma_{\kappa}}.
    \label{eqn:kappa_norm} 
\end{align}


Subsequently, the normalized values are used to calculate concentration weights using the formula below:
\begin{equation}
\label{eqn:kappa_w}
w_{c}^{k} = f_{k}(\Tilde{\kappa_{c}}) = 1 - \sigma(\Tilde{\kappa_{c}} \times T),
\end{equation}
where $\sigma$ is a well-known sigmoid function, and $\emph{T} \in (0, 1)$ is a hyper-parameter that denotes a temperature value.  \\

\begin{table*}[!t]
\begin{center}
\caption{Verification performance, in terms of accuracy, results (\%) on LFW, YTF, two pose benchmarks (CFP-FP and CPLFW), and two age benchmarks (AgeDB and CALFW). }
\label{tb:main_exp}
\resizebox{.82\textwidth}{!}{%
    \begin{tabular}{ l|c|c|c|c|c|c} 
    \toprule
    Method & LFW & CFP-FP & CPLFW & AgeDB & CALFW & YTF\\
    \hline \hline
    Center Loss  \cite{wen2016discriminative} & 99.27 & - & 81.40 & - & 90.30 & 94.9\\
    SphereFace \cite{liu2017sphereface} & 99.27 & - & 81.40 & - & 90.30 &95.0\\
    VGGFace2  \cite{VGGFace2} & 99.43 & - & 84.00 & - & 90.57 &-\\ 
    UV-GAN  \cite{deng2018uv} & 99.60 & 94.05 & - & - & - & -\\
    ArcFace  \cite{deng2019arcface} & \underline{99.82} & 98.27 & 92.08 & 98.15 & 95.45 & 98.0\\
    
    ArcFace-SCF   \cite{li2021spherical} & \underline{99.82} & 98.40 & 93.16 & 98.30 & 96.12 & -\\
    CurricularFace  \cite{huang2020curricularface}  & 99.80 & 98.37 & 93.13 & 98.32 & \underline{96.20} & -\\
    MagFace \cite{meng2021magface} & \textbf{99.83} & 98.46 & 92.87 & 98.17 & 96.15 & -\\
    
    \hline
    KappaFace (\textbf{\texttt{Memory buffer}}) & \textbf{99.83} & \textbf{98.69} & \underline{93.22} & \textbf{98.47} & \textbf{96.23}  & \textbf{98.0}\\
        KappaFace (\textbf{\texttt{Momentum encoder}}) & \textbf{99.83} & \underline{98.60} & \textbf{93.40} & \underline{98.35} & {96.15}  & \textbf{98.0}\\
    \bottomrule
    \end{tabular}
    }
\end{center}
\end{table*}

\textbf{Note.} The smaller the temperature value, the closer the weights are scattered around 0.5, $\sigma(0) = 0.5$. Also, note that the greater the concentration value than is the mean, the smaller the concentration weight $w_{c}^{k}$, and vice versa. Here, the memory buffer or the momentum encoder plays a crucial role in computing all the concentration values. Thus, we can encourage the concentration values to be closer to the mean and less diverse. Moreover, the class learning difficulty is determined by the relative concentration $\Tilde{\kappa_{c}}$ over the entire training dataset instead of its true value.

In contrast, considering that the number of samples per class is fixed throughout the training process, there is no need to update the population weights once they are computed. Thus, the population weights are obtained as follows:
\begin{equation}
\label{eqn:sample_w}
w_{c}^{s} = f_{s}(n_c) = \frac{\cos(\pi \times \frac{n_{c}}{K}) + 1}{2},
\end{equation}
where $n_{c}$ is the number of samples for the \emph{c}-th class, and \emph{K} is chosen to be the maximum among them. {Unlike  {the concentration weights} $w^{k}_{c}$,  {the sample weights} $w_{c}^{s}$ are constant throughout the training. Hence, they do not need updates. One may replace them with a constant value for all classes; however, our idea is to prioritize underrepresented classes in case of values for $w^{k}_{c}$ are equal. Therefore, we propose different constants for each class depending on how represented they are.} 

Finally, the calibration parameter for the margin value of instances belonging to the c-\textit{th} class is determined as follows:
\begin{equation}
\label{eqn:final_margin}
\psi_{c} = (1-\gamma) w_{c}^{s} + \gamma w_{c}^{k},
\end{equation}
 where $\gamma \in (0,1)$ is a hyper-parameter that balances the contribution of the concentration and population weights. We apply the additive margin values proposed by ArcFace. Thus, $m_0$ is a fixed initial margin value and the final classification loss can be defined as follows:
\begin{equation}
\label{eqn:kappa_loss}
\mathcal{L_K} =   -\log \frac{e^{s(\cos(\theta_{y_{i}} + \psi_{y_i} \cdot m_{0}))}}{e^{s(\cos(\theta_{y_{i}} + \psi_{y_i} \cdot m_{0}))} + \sum_{j=1, j\neq {y}_{i} }^{C}e^{s(\cos(\theta_{j}))}}.
\end{equation}
Therefore, using Eq.~\ref{eqn:kappa_loss}, we can dynamically adjust the angular margin based on the class learning difficulty and its population.\\

\subsection{Gradient Analysis of KappaFace }
In order to provide a comparison with ArcFace, we first define the following  to measure the effectiveness of our KappaFace in the training phase. We demonstrate the effectiveness using the gradient analysis as follows:

\textbf{Theorem 1.} \emph{Given $\{x_i, y_i\}_{y_i=l}$, the softmax loss in Eq. \ref{eqn:kappa_loss} provides a smaller correction on $\mathbf{W}$ than ArcFace loss provided samples of a class $l$ have a high concentration, and vice versa.}

\textit{Proof: }From Eq. \ref{eqn:kappa_loss}, we derive the gradient of our KappaFace loss for one sample $\{z_i, y_i\}$, $y_i=l \in \{1,...,C\}$  with respect to each $W_k$ as follows:
\begin{align}
\label{eqn:derivative}
  \nabla_{\mathbf{W}_k} \mathcal{L_K} =\left\{
  \begin{array}{@{}ll@{}}
    - s (1-p_{k})\sin (\theta _{k} +\psi_{l} m_0)\cdot\nabla_{\mathbf{W}_k} \theta_{k}, & \ k=l \\
    s  p_{k} \sin \theta_{k} \cdot \nabla_{\mathbf{W}_k} \theta_{k}, & \ k \neq l
  \end{array}\right.
\end{align} 
where 
\begin{equation}
\label{eqn:probability}
    p_k = \frac{\exp (s \cos (\theta _{k} +\mathbbm{1}_{l} \cdot \psi_{l} m_0 ))}{\exp (s\cos (\theta _{k} + \psi_{l} m_0)) +\sum_{j=1, j\neq l}^{C}\exp(s\cos \theta_{j})}.
\end{equation}

For convenience, let $\mathring{p}_{k}$ and $\mathring{m}$ be the probabilities and the fixed additive margin, respectively, corresponding to the ArcFace loss. Then, if $l$ is a high concentration class, the scaler $\psi_{l}$ modulates its margin to be smaller than fixed $\mathring{m}$. From Eq. \ref{eqn:probability} and $\sum_k^C p_k = \sum_k^C \mathring{p}_k=1$, we have:
\begin{equation}
\label{eqn:cons_prop}
     \begin{cases}
                    p_{l} > \mathring{p}_{l} \\
                    p_{k} < \mathring{p}_{k} &\text{, $\forall k \neq l$}
                    \end{cases} .
\end{equation}

In addition, considering the \textbf{first case} in Eq. \ref{eqn:derivative}, when  $k=l$ and $\theta_l$ is small enough, we also obtain the following:
\begin{equation}
\sin (\theta_{l} +  \psi_{l} m_{0}) < \sin (\theta_{l} + \mathring{m}_{l}).
\end{equation}
Thus, combining with the \textbf{first property} in Eq. \ref{eqn:cons_prop}, we show that $\Vert \nabla_{\mathbf{W}_l} \mathcal{L_K} \Vert$ of~\SystemName~loss  is  smaller than that of ArcFace loss.

In the {second case} of Eq. \ref{eqn:derivative}, $k \neq l$, as $p_{k} < \mathring{p}_{k} \textrm{  }  \forall k \neq l$ (second property in Eq. \ref{eqn:cons_prop}), we achieve the same property of $\nabla_{\mathbf{W}_k} \mathcal{L_K}$ as the ArcFace loss. In both cases above, we can inversely deduce the outcomes for the cases of $l$ being a low concentration class.

Finally, we summarize our end-to-end training process with our proposed~\SystemName~loss in Algorithm \ref{alg:kappaface}. In the inference phase, the memory buffer or the momentum encoder is removed so that the computation expense of our proposed method is similar to the ArcFace.

%% file: Sections/4_experiment.tex
\begin{table}[t!]
\caption{{1:1 verification TAR, measured in \%, (@FAR= $1e^{-4}$) on the IJB-B and IJB-C datasets, where the best performances are shown in bold, and the second best are underlined.}}
\label{tb:ijb_exp}
\begin{center}
\resizebox{.49\textwidth}{!}{%
    \begin{tabular}{ l|c|c} 
    \toprule
    Method & IJB-B & IJB-C\\
    \hline \hline
    VGGFace2  \cite{VGGFace2} & 80.0 & 84.1 \\
    Multicolumn  \cite{xie2018multicolumn} & 83.1 & 86.2\\
    P2SGrad  \cite{zhang2019p2sgrad} & - & 92.3\\
    Adacos  \cite{zhang2019adacos} & - & 92.4\\
    ArcFace-VGG2-R50  \cite{deng2019arcface} & 89.8 & 92.1\\
    ArcFace-MS1MV2-R100  \cite{deng2019arcface} & 94.2 & 95.6\\
   {CurricularFace-MS1MV2-R100}  \cite{huang2020curricularface}  & {94.8} & {96.1}\\
    {MagFace-MS1MV2-R100}  \cite{meng2021magface}  & {94.5} & {96.0}\\
    \hline
    KappaFace-MS1MV2-R100 (\textbf{\texttt{Memory buffer}}) & \underline{95.1} & \underline{96.4} \\
        KappaFace-MS1MV2-R100 (\textbf{\texttt{Momentum encoder}}) & \textbf{95.3} & \textbf{96.6} \\
    \bottomrule
    \end{tabular}%
}
\end{center}
\end{table}

\begin{table}[th!]
\begin{center}
\caption{{Verification performance, in terms of accuracy, compared with SOTA methods on MegaFace Challenge 1 using FaceScrub~\cite{ng2014data} as the probe set. The reported results are based on the refined data. Id corresponds to the rank 1 face identification accuracy with 1M distractors, whereas Ver refers to the face verification TAR (@FAR = $1e^{-6}$). The best performances are shown in bold, and the second best are underlined.}}
\label{tb:mf_exp}
\resizebox{.49\textwidth}{!}{%
    \begin{tabular}{ l|c|c} 
    \toprule
    Method & Id & Ver\\
    \hline \hline
    AdaptiveFace \cite{liu2019adaptiveface} & 95.02 & 95.61\\
    P2SGrad  \cite{zhang2019p2sgrad} & 97.25 & -\\
    Adacos  \cite{zhang2019adacos} & 97.41 & -\\
    CosFace \cite{wang2018cosface} & 97.91 & 97.91\\
    MV-AM-Softmax-a \cite{wang2020mis} & 98.00 & 98.31\\
    ArcFace-MS1MV2-R100  \cite{deng2019arcface} & 98.35 & 98.48\\
    
    {CurricularFace-MS1MV2-R100}  \cite{huang2020curricularface}  & {98.71} &  {98.64}\\
    {MagFace-MS1MV2-R100}  \cite{meng2021magface}  & {97.78} & {98.10}\\
    \hline
    KappaFace-MS1MV2-R100 (\textbf{\texttt{Memory buffer}}) & \underline{98.77} & \textbf{98.91} \\
        KappaFace-MS1MV2-R100 (\textbf{\texttt{Momentum encoder}}) & \textbf{98.78} & \underline{98.83} \\
    \bottomrule
    \end{tabular}%
}
\end{center}
\end{table}
\section{Experimental results}
\label{sec:exp_results}
{In this section, we introduce nine facial benchmark datasets used to validate our proposed Kappaface. Subsequently, we compare the performance of our method with state-of-the-art (SoTA) baselines across these datasets. Lastly, we conduct ablation studies to highlight the contributions of hyper-parameters (in our proposed method) individually.}

\subsection{Implementation Details}
\textbf{Datasets.} We utilize the MS1MV2   \cite{deng2019arcface} as our training set to ensure a fair comparison with other prior state-of-the-art methods, where MS1MV2 contains approximately 5.8M images of 85,742 identities. In addition, we perform computationally extensive experiments on several popular benchmarks, namely LFW \cite{huang2008labeled}, CFP-FP   \cite{sengupta2016frontal}, CPLFW   \cite{zheng2018cross}, AgeDB   \cite{moschoglou2017agedb}, CALFW   \cite{zheng2017cross}, YTF \cite{whitelam2017iarpa}, IJB-B   \cite{whitelam2017iarpa}, IJB-C   \cite{maze2018iarpa}, and MegaFace Challenge 1 \cite{kemelmacher2016megaface}.

\begin{figure}[t!]
\centering
\includegraphics[width=8cm]{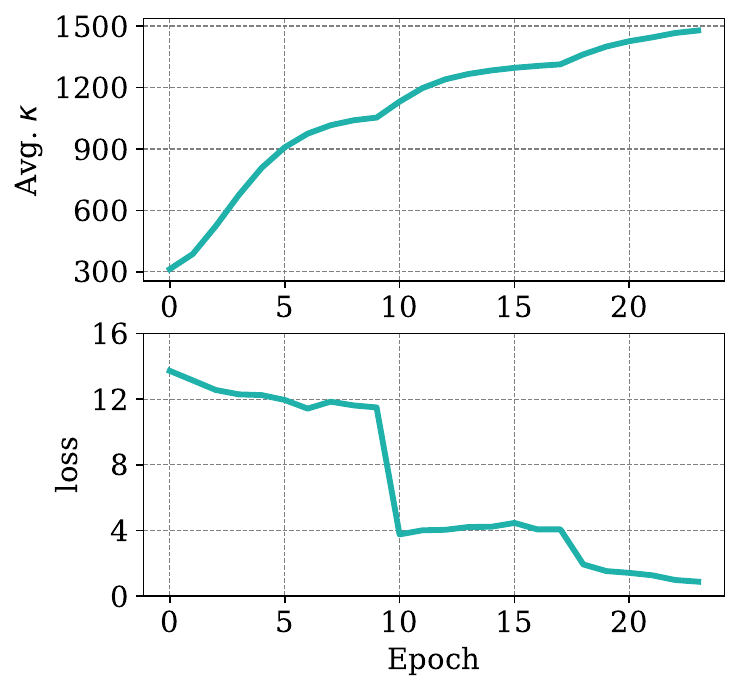}
\caption{Training convergence of our proposed method depicted through the margin loss values (bottom) and the average concentration (top) across training epochs.} 
\label{fig:training_loss}
\end{figure}
\textbf{Training Setting.} For data pre-processing, we follow  ArcFace \cite{deng2019arcface} and obtain the normalized face crops (112 × 112) by utilizing five landmarks \cite{zhang2016joint, tai2019towards}. For the embedding network, we adopt ResNet100 as reported in   \cite{deng2019arcface}. We train models on NVIDIA GeForce RTX 3090 Ti with a batch size of 512. The models are trained using the Stochastic Gradient Descent (SGD) algorithm, with the momentum parameter of 0.9 and the weight decay of $5e - 4$. We divide the learning rate  {by 10} at the 10th, 18th, and 22nd epochs, and finish at 24th epochs. Regarding the memory buffer, we set the momentum parameters $\alpha = 0.3$ and $m=0.999$ respectively, whereas $\gamma$ is set to 0.7. Also, we fix the initial margin value for~\SystemName~(\textbf{\texttt{Momentum encoder}}) for a couple of epochs to avoid {convergence failure}. All experiments are implemented using PyTorch \cite{paszke2019pytorch, insightface}. Additionally, we utilize Partial FC \cite{an2020partical_fc}, which is a sparse variant of the model parallel architecture for large scale face recognition.

{Our training convergence, with the memory buffer approach, is illustrated in Fig. {\ref{fig:training_loss}}. In addition, we also include the average concentration parameters across the training epochs. As we can observe, as the training loss decreases, the average concentration increases, demonstrating the convergence capabilities of our algorithm.}

{\textbf{Testing.} We adopt the evaluation setting from ~\mbox{\cite{wang2018cosface, liu2017sphereface}}. A representation for a test sample is obtained by concatenating the embeddings of the original image with that of its horizontally flipped view. For the verification task on LFW, CFP-FP, CPLFW, AgeDB, CALFW, and YTF datasets, assume there are $n$ image pairs, consisting of $n_p$ positive pairs and $n_n$ negative pairs. The performance of the method is reported using the accuracy metric (\%) at the optimal validation threshold $t$. This is given by:}
\begin{equation}
Acc = \frac{\Sigma_{i=1}^{n_p}\mathbb{I}(\cos(p_i)\leq t) + \Sigma_{i=1}^{n_n}\mathbb{I}(\cos(p_i) > t)}{n}\times 100,
\end{equation}
{where $\cos(p_i)$ represents the cosine similarity of the $i$-th image pair.

For the IJB-B/C datasets, we employ the True Acceptance Rate (TAR) as a performance metric. TAR measures the proportion of genuine matches that the system correctly accepts. Alongside TAR, we use the False Acceptance Rate (FAR) to quantify the likelihood that the system incorrectly recognizes a mismatch as a legitimate match. These metrics are mathematically expressed as:}
\begin{align}
\text{TAR} & = \frac{\Sigma_{i=1}^{n_p}\mathbb{I}(\cos(p_i)\leq t)}{n_p}\times 100,\\
\text{FAR} & = \frac{\Sigma_{i=1}^{n_n}\mathbb{I}(\cos(p_i)\leq t)}{n_n}\times 100.
\end{align}
{Typically, TAR is reported at a specific FAR level. Following the common settings in ~\mbox{\cite{deng2019arcface}}, we evaluate using TAR@FAR$=1e^{-4}$, TAR@FAR$=1e^{-6}$   for IJB-B/C and MegaFace datasets, respectively.

For the identification task on the MegaFace dataset, we utilize rank-1 accuracy as our evaluation metric. Given 
$n_g$ images in the gallery set and  $n_p$  images in the probe set, the formula for rank-1 accuracy is:}
\begin{equation}
    Acc@1 = \frac{\Sigma_{i=1}^{n_p}\mathbb{I}(y_i=lb(\text{Neighbor}(\text{img}_i))}{n_p}\times 100,
\end{equation}
{where $lb(\text{Neighbor}(\text{img}_i))$  represents the label of the nearest neighbor of $\text{img}_i$ within the  $n_g$  gallery images.}

{For comparison, we select recent state-of-the-art (SOTA) methods that are closely related to ours. These include Center Loss~\mbox{\cite{wen2016discriminative}},
SphereFace~\mbox{\cite{liu2017sphereface}},
 VGGFace2~\mbox{\cite{VGGFace2}}, 
UV-GAN~\mbox{\cite{deng2018uv}},
 Multicolumn~\mbox{\cite{xie2018multicolumn}}, AdaptiveFace~\mbox{\cite{liu2019adaptiveface}},
P2SGrad~\mbox{\cite{zhang2019p2sgrad}},
 Adacos~\mbox{\cite{zhang2019adacos}}, 
 CosFace~\mbox{\cite{wang2018cosface}},
MV-AM-Softmax-a~\mbox{\cite{wang2020mis}},
 ArcFace~\mbox{\cite{deng2019arcface}},
CurricularFace~\mbox{\cite{huang2020curricularface}},
MagFace~\mbox{\cite{meng2021magface}}. In each experiment, we only include methods that have previously reported their performance. Additionally, we compare with probabilistic models such as  PFE-G~\mbox{\cite{shi2019probabilistic}}, SCF~\mbox{\cite{li2021spherical}}, and LDAM-DRW~\mbox{\cite{cao2019learning}}}.
\subsection{Evaluation Results}

\textbf{Results on LFW, CFP-FP, CPLFW, AgeDB,
CALFW, and YTF.} We train our KappaFace on MS1MV2 using ResNet100 as a backbone and provide a comparison with other SOTA methods. In this paper, we similarly follow the unrestricted with labeled outside data protocol, as performed in \cite{deng2019arcface, huang2020curricularface}, to report the results. Additionally, we used 7,000 pair testing images for CFP-FP and 6,000 pairs for the rest of the datasets to report our results similar to ArcFace\cite{deng2019arcface}. As shown in Table \ref{tb:main_exp}, our method achieves better results on LFW, YTF and CALFW, where the performance is almost saturated. However, our KappaFace demonstrates a superiority by outperforming counterparts by a significant margin on most of the other datasets, such as CFP-FP, CPLFW and AgeDB, which can be clearly observed from the result table. {The performance on LFW is saturated; however, we still include this benchmark for consistency with previous works. For the current SOTA methods, the LFW dataset is fairly easy, since there is not much variation in terms of identity's facial angles and ages.} 
Similarly, our proposed method achieves higher accuracy than ArcFace-SCF \cite{li2021spherical}, which is the SOTA method that also inherits the spherical distribution properties. However, it is applied on individual samples instead. 
\begin{figure}[t!]
    \centering
    \subfloat[ROC for IJB-B ]{{\includegraphics[width=8.2cm]{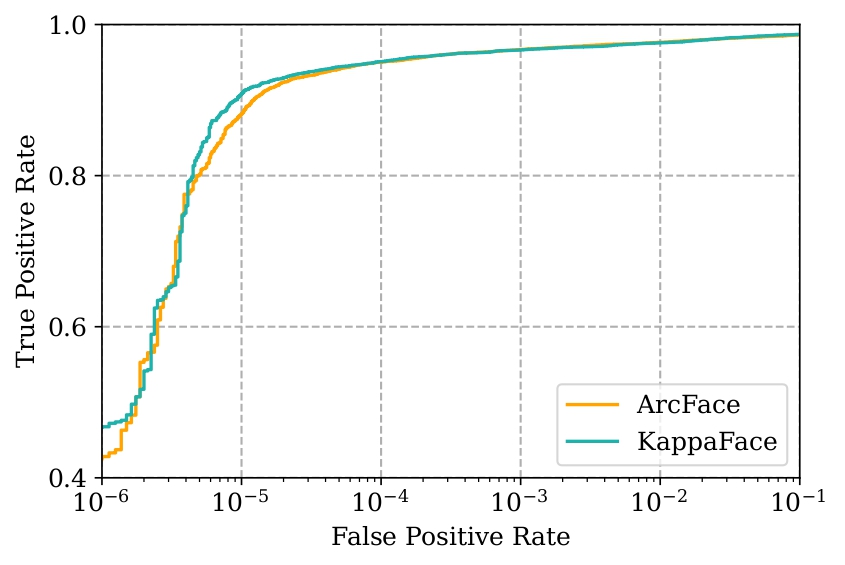} }}
    \qquad
    \subfloat[ROC for IJB-C]{{\includegraphics[width=8.2cm]{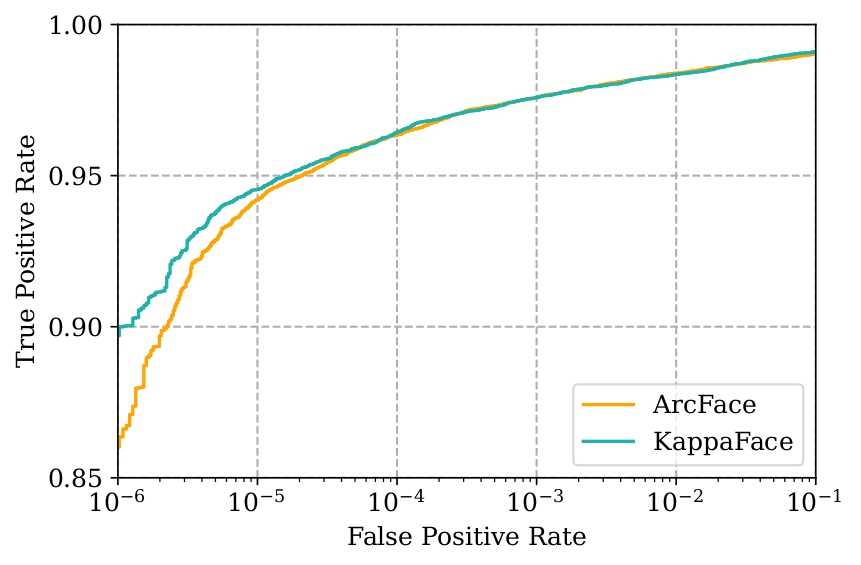} }}%
    \caption{\textbf{ROC of 1:1 verification protocol on IJB-B and IJB-C}. A comparison of KappaFace
(Memory buffer) and ArcFace on IJB-B and IJB-C, where
our KappaFace demonstrates a superiority over ArcFace even at FAR=$1e^{-6}$.}%
    \label{fig:roc-ijb}%
\end{figure}

\textbf{Results on IJB-B and IJB-C.} The IJB-B dataset consists of 1,845 identities having 21.8K still images and 55K frames collected from 7,011 videos. There are 10,270 positive pairs and 8M negative pairs used for the 1:1 verification. IJB-C is an extension of IJB-B that includes 3,500 subjects with 31K images and 117K frames. There are 19,557 positive pairs and 15,638,932 negative pairs used for the 1:1 verification. Finally, we use our pre-trained KappaFace-MS1MV2-R100 and validate the model on the IJB-B and IJB-C datasets. In Table \ref{tb:ijb_exp}, we provide a comparison in terms of True Acceptance Rate (TAR) (@FAR=$1e^{-4}$) with other state-of-the-art methods. Table~\ref{tb:ijb_exp} demonstrates that KappaFace with the momentum encoder clearly boosts the performance on both datasets by achieving the accuracy of $95.3\%$ and $96.6\%$ on IJB-B and IJB-C, respectively. In Fig. \ref{fig:roc-ijb}, we present the full ROC curves comparing our KappaFace (\textbf{\texttt{Memory buffer}}) and ArcFace on IJB-B and IJB-C, where our KappaFace demonstrates a staggering improvement in performance over ArcFace even at FAR=$1e^{-6}$ establishing a new baseline performance.


\begin{table}[t!]
\caption{Ablation study on verification performance with ResNet34 + Momentum Encoder on LFW, CFP-FP, AgeDB, CALFW and CPLFW.}
\label{tb:small_bb_exp}
\resizebox{.495\textwidth}{!}{%
    \begin{tabular}{ l|c|c|c|c|c} 
    \toprule
    Method & LFW & CFP-FP & CPLFW & AgeDB & CALFW \\
    \hline \hline
    ArcFace  \cite{deng2019arcface} & 99.68 & 94.04 & 89.30 &  96.60 & 95.00\\
    ArcFace-PFE-G   \cite{shi2019probabilistic} & 99.71 & 94.19 & 89.90 & 96.70 & 95.49 \\
    ArcFace-SCF   \cite{li2021spherical} & 99.73 & 94.87 & 90.62 & 97.16 & 95.79 \\
    {LDAM-DRW} \cite{cao2019learning} & {99.80} & {97.41} &{91.73} & {97.72} & {95.95}\\
    \hline
        KappaFace  & \textbf{99.82} & \textbf{97.57} & \textbf{92.00} & \textbf{98.02} & \textbf{96.13}\\
    \bottomrule
    \end{tabular}
    }
\end{table}

\begin{table}[t!]
\begin{center}
\caption{{Ablation study on $\gamma$ with ResNet34 + Momentum Encoder on LFW, CFP-FP, AgeDB, CALFW and CPLFW.}}
\label{tb:ablate_gamma}
\resizebox{.46\textwidth}{!}{%
    \begin{tabular}{ l|c|c|c|c|c } 
    \toprule
    $\gamma$ & LFW & CFP-FP & CPLFW & AgeDB & CALFW  \\
    \hline \hline
    ArcFace  \cite{deng2019arcface} & 99.68 & 94.04 & 89.30 &  96.60 & 95.00 \\
    \hline
    0.0   & {99.82} & {96.29} & {91.13} & \textbf{98.03} &{96.07}\\
    0.3    & {99.82} & {96.72} & {91.71} & {97.95} & {96.07}\\
    0.5    & {99.82} & {97.04} & {91.53} & {97.85} & {96.05}\\
    0.7  & \textbf{99.82} & \textbf{97.57} & {92.00} & {98.02} & \textbf{96.13} \\
    1.0  & {99.80} & {97.57} & \textbf{92.10} & {97.97} & {96.12} \\
    \bottomrule
    \end{tabular}
    }
\end{center}
\end{table}

\textbf{Results on MegaFace.} Next, we provide an evaluation result on the MegaFace Challenge 1. The gallery set of MegaFace consists of 1M images of 690K unique individuals, and the probe set has 100K photos of 530 subjects from FaceScrub. We report the results under a large training set protocol, where we train ResNet100 on MS1MV2. Note that the results reported are based on the refined MegaFace, where noisy labels are removed. Table \ref{tb:mf_exp} clearly shows that our method outperforms other state-of-the-art methods in both identification and verification tasks, where \SystemName{} trained with the memory buffer achieves 98.77 and 98.91, respectively.

\textbf{Probabilistic models with smaller backbone. } In this study, we compare our approach with SOTA probabilistic deep face models, namely ArcFace-PFE-G \cite{shi2019probabilistic}, ArcFace-SCF \cite{li2021spherical}, and {LDAM-DRW}~\cite{cao2019learning},  {with a smaller backbone (ResNet34). We note that LDAM-DRW is proposed to address imbalanced datasets in classification problems; however, as it is closely related to our work, we fine tuned it for the deep face embedding model.} As shown in Table \ref{tb:small_bb_exp}, all three probabilistic models present a considerable improvement upon the ArcFace baseline across all the experimental datasets. Our \SystemName, on the other hand, consistently outperforms the counterparts by a decent margin. Notably, it achieves a staggering $99.82 \%$ accuracy on LFW dataset, which is comparable to the results obtained from ResNet100.
\begin{table}[t!]
\begin{center}
\caption{{Ablation studies on our hyperparameters with ResNet100 + Momentum Encoder, where the performance gains are shown in green, and drops in red compared to the optimal setting (shown in bold).}}
\label{tb:ablation}
\resizebox{.49\textwidth}{!}{%
    \begin{tabular}{ c|c|c|c|c|c|c } 
    \toprule
    \multicolumn{2}{c|}{Hyper-params}  & \multicolumn{5}{c}{Datasets} \Bstrut\\ 
    \hline
    $m_0$ & $T$ & LFW & CFP-FP & AgeDB & IJB-B & IJB-C \Tstrut\\
    \hline \hline
    0.8 & 0.4 & \textbf{99.83} & \textbf{98.60} & \textbf{98.35} & \textbf{95.3} & \textbf{96.6}\Tstrut\\
    \hline
    0.8 & \textit{0.3} & 99.83 & 98.63 \textcolor{green}{(+0.03)} &  98.32 \textcolor{red}{(-0.03)} & 95.17 \textcolor{red}{(-0.13)} & 96.44 \textcolor{red}{(-0.16)}\Tstrut\\
    0.8 & \textit{0.5} & 99.83 & 98.63 \textcolor{green}{(+0.03)} & 98.35 &  95.29  \textcolor{red}{(-0.01)} & 96.53 \textcolor{red}{(-0.07)}\\
    \hline
    \textit{0.7} & 0.4  & 99.83 & 98.71 \textcolor{green}{(+0.11)} & 98.40 \textcolor{green}{(+0.05)} &  95.10 \textcolor{red}{(-0.20)} & 96.53  \textcolor{red}{(-0.07)}\Tstrut\\
    \textit{0.9} & 0.4 & 99.83 & 98.51 \textcolor{red}{(-0.09)} & 98.35 & 95.24 \textcolor{red}{(-0.06)} & 96.50 \textcolor{red}{(-0.10)}\Bstrut\\
    \bottomrule
    \end{tabular}%
}
\end{center}
\end{table}

\begin{figure*}[t!]
\centering
\includegraphics[width=6.6in]{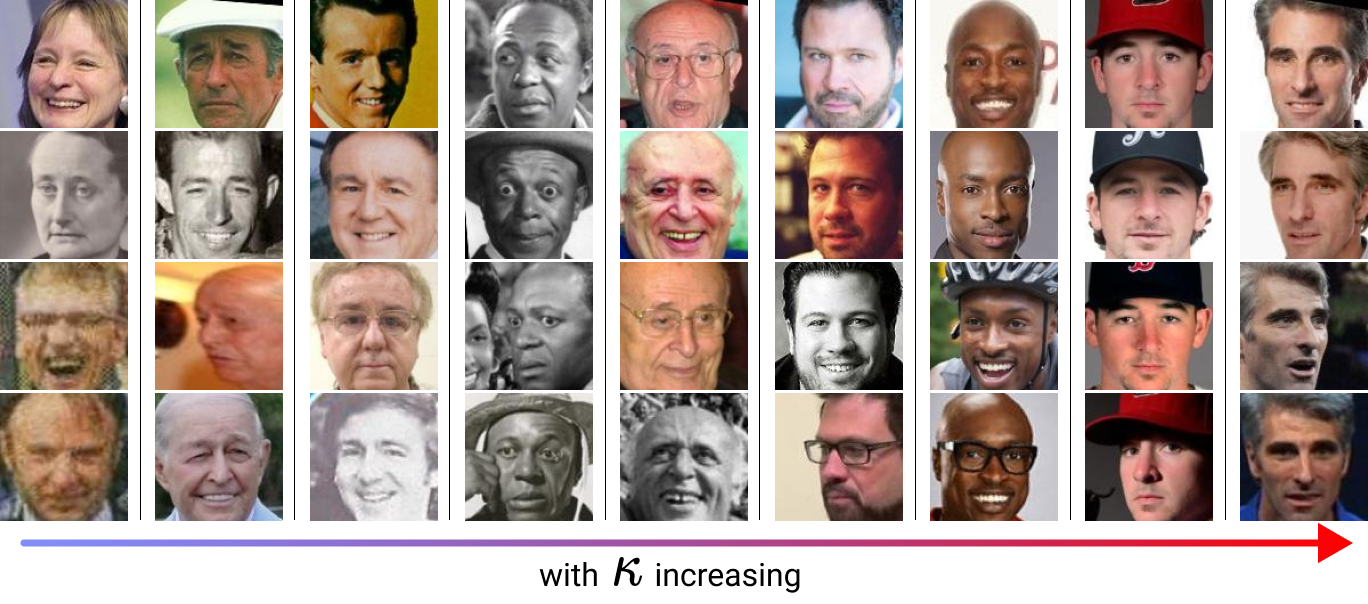}
\caption{\textbf{Identity vs. concentration value.} Each column corresponds to an identity, with 4 random samples provided, and their $\kappa$s are sorted in ascending order from left to right. {We demonstrate that $\kappa$ can be affected by how noisy, diverse or available samples are.}} 
\label{fig:kappa_illustration}
\end{figure*}
\begin{figure*}[t!]
\centering
\includegraphics[width=6.8in]{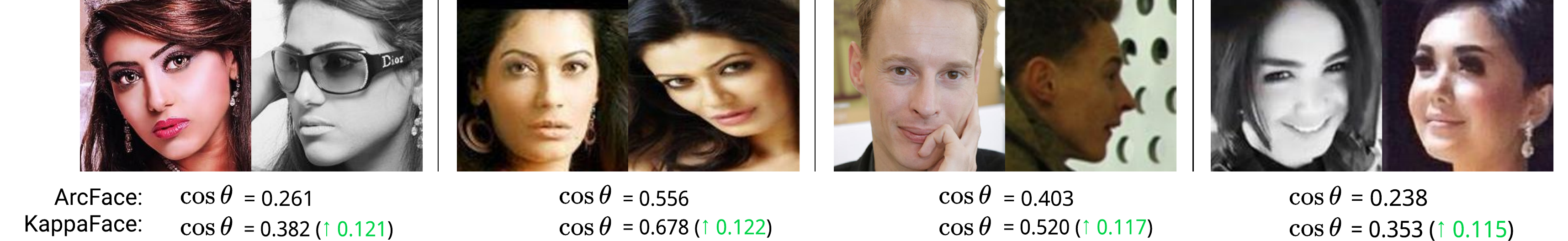}
\caption{\textbf{Positive pairs from IJB-C dataset.} $\cos \theta$ is the cosine distance of a verification pair $x_1$ and $x_2$. The lower the $\cos \theta$, the more likely that the model has larger false positive rate. Thus, the green values indicate the improvements over ArcFace. } 
\label{fig:pos_pairs}
\end{figure*}

\begin{figure*}[t!]
\centering
\includegraphics[width=6.8in]{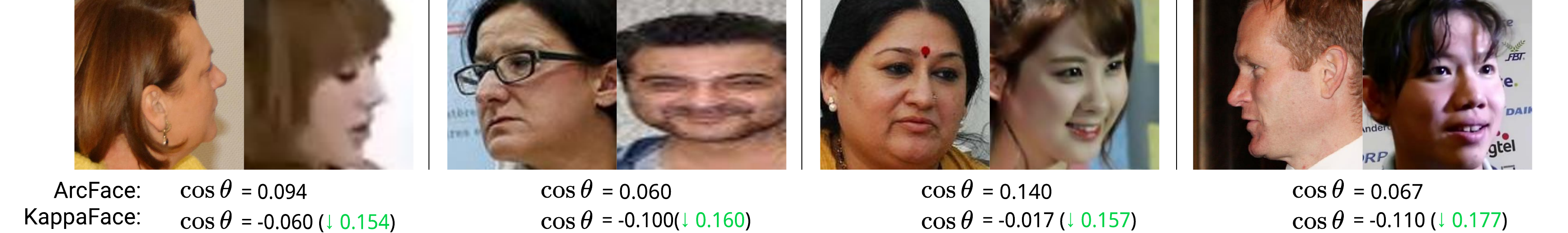}
\caption{\textbf{Negative pairs from IJB-C dataset.} $\cos \theta$ is the cosine distance of a verification pair $x_1$ and $x_2$. The greater the $\cos \theta$, the more likely that the model has larger false negative rate. Thus, the green values indicate the improvements over ArcFace.} 
\label{fig:neg_pairs}
\vspace{-8pt}
\end{figure*}

 
\textbf{Qualitative analysis}. We provide Figure \ref{fig:kappa_illustration} to demonstrate that the $\kappa$ value can represent a class learning difficulty in the training dataset, MS1MV2. The class samples in the left-most column are noisy and have different qualities; hence its $\kappa$ is the smallest. Moving from the left to right-hand side column, identities vary less in terms of age and the sample quality gradually improves, indicated by the increasing $\kappa$. Finally, we also perform an ablation study on the IJB-C dataset to show the advantages of our proposed \SystemName{} over the baseline ArcFace model in the verification task. As shown in Figure \ref{fig:pos_pairs} and Figure \ref{fig:neg_pairs}, our method, \SystemName, efficiently represents the positive images with higher cosine similarity values compared to {the} ArcFace. Meanwhile, for the negative pairs, the \SystemName{} can push their representations away farther apart, showing through the smaller similarity values. 


{\textbf{Contribution of class concentration and population weights}. We ablate along the hyper-parameter $\gamma$ with MS1MV2 dataset and ResNet34 backbone. From the results in Table} \ref{tb:ablate_gamma}, {we note that at any values of $\gamma$ our model can substantially outperform the baseline ArcFace} \cite{deng2019arcface}{. Moreover, with $\gamma=0.7$, as we suggested, }\SystemName{}{ obtains the optimal performance.} {And, we also show that both terms $w_{c}^{k}$ and $w_{c}^{s}$ contribute to our final results. Note that only $w_{c}^{k}$ changes throughout the training, hence one may replace or drop $w_{c}^{s}$. However, we believe it encourages to learn under-represented categories in case concentration values for well-represented and poorly-represented identities are equal.}

\textbf{Hyper-parameter fine-tuning. }In this study, we use ResNet100 with the momentum encoder to analyze how various settings of $T$ and $m_0$ affect ~\SystemName~ on the MS1MV2. We select LFW, CFP-FP, AgeDB, IJB-B and IJB-C datasets for our experiments. 
As we can examine from  Table \ref{tb:ablation}, the optimal setting observed in our series of experiments was $m_{0} = 0.8$, and $T = 0.4$. However, we would also like to emphasize that some other settings perform better on individual datasets. For example, ~\SystemName~ with  $m_{0} = 0.7$ and $T = 0.4$ works better on CFP-FP and AgeDB yet we observe the performance drop on IJB-B and IJB-C datasets. 
Our proposed ~\SystemName~ shows the robustness and improvements over the fixed margin used by the ArcFace loss.  Table ~\ref{tb:ablation} demonstrates that our proposed method outperforms ArcFace in all the selected evaluation datasets. 

{\textbf{Discussion \& Limitations.}
From our experiments, we have demonstrated that the advancements we have made provide a significant contribution to the deep face recognition landscape. This progress is evident not only when our approach is compared to prior methods but also when set against alternative techniques. Importantly, deep face recognition plays a pivotal role in shaping critical verification systems for real-world applications. Therefore, even minor improvements can represent a substantial advancement in enhancing security and driving the field's evolution. Beyond quantitative improvements, our method offers qualitative value by introducing a streamlined dynamic adjustment mechanism for margin losses, potentially paving the way for future breakthroughs in the domain.
}

{While our method addresses the issue of imbalance in face recognition datasets, we acknowledge its dependence on an auxiliary model or a memory buffer to determine the concentration value for each class; however, they bring insignificant computational overhead. Furthermore, we argue that beyond the additive angular margin loss, our proposed technique could be adapted to various other margin-based losses with the right design. We leave this as an opportunity for future investigations.}

%% file: Sections/5_conclusion.tex
\section{Conclusion}
\label{sec:conclusion}
In this paper, we address the challenge of deep face recognition when confronted with highly imbalanced datasets and introduce a novel Adaptive Additive Margin Loss, termed KappaFace, that dynamically adjusts the positive additive margins based on class imbalance and associated learning difficulties. Our primary contribution lies in reducing the margins for classes that are easily learned and display high concentration values, while simultaneously increasing the margins for the counter-classes by leveraging the von Mises-Fisher (vMF) distribution. Additionally, we verify the generalizability of our proposed KappaFace via two approaches for modeling hyperspherical distribution in the latent space, namely the momentum encoder and the memory buffer. These approaches neither introduce additional computational costs nor incur memory expenses during the inference phase. {Our extensive experiments across nine well-known facial benchmarks, tested against eight contemporary baselines, demonstrate that KappaFace consistently outperforms other state-of-the-art methods, displaying an improvement of up to 0.5\% on the IJB-B/C datasets. Moreover, our adaptive method showcases impressive generalization capabilities, especially with highly imbalanced datasets.}

{While our method exhibits evident improvements throughout the evaluation experiments, it still relies on an auxiliary model or memory module during training. Additionally, its current configuration is primarily tailored to adaptively adjust the angular margin built on top of the ArcFace method. Consequently, our future research aims to estimate the class learning difficulties more efficiently, while ensuring compatibility with a wider range of similar algorithms.}



%% file: MAIN.bbl
\begin{thebibliography}{10}
\providecommand{\url}[1]{#1}
\csname url@samestyle\endcsname
\providecommand{\newblock}{\relax}
\providecommand{\bibinfo}[2]{#2}
\providecommand{\BIBentrySTDinterwordspacing}{\spaceskip=0pt\relax}
\providecommand{\BIBentryALTinterwordstretchfactor}{4}
\providecommand{\BIBentryALTinterwordspacing}{\spaceskip=\fontdimen2\font plus
\BIBentryALTinterwordstretchfactor\fontdimen3\font minus
  \fontdimen4\font\relax}
\providecommand{\BIBforeignlanguage}[2]{{%
\expandafter\ifx\csname l@#1\endcsname\relax
\typeout{** WARNING: IEEEtran.bst: No hyphenation pattern has been}%
\typeout{** loaded for the language `#1'. Using the pattern for}%
\typeout{** the default language instead.}%
\else
\language=\csname l@#1\endcsname
\fi
#2}}
\providecommand{\BIBdecl}{\relax}
\BIBdecl

\bibitem{wang2017normface}
F.~Wang, X.~Xiang, J.~Cheng, and A.~L. Yuille, ``Normface: L2 hypersphere
  embedding for face verification,'' in \emph{Proceedings of the 25th ACM
  international conference on Multimedia}, 2017, pp. 1041--1049.

\bibitem{schroff2015facenet}
F.~Schroff, D.~Kalenichenko, and J.~Philbin, ``Facenet: A unified embedding for
  face recognition and clustering,'' in \emph{Proceedings of the IEEE
  conference on computer vision and pattern recognition}, 2015, pp. 815--823.

\bibitem{wen2016discriminative}
Y.~Wen, K.~Zhang, Z.~Li, and Y.~Qiao, ``A discriminative feature learning
  approach for deep face recognition,'' in \emph{European conference on
  computer vision}.\hskip 1em plus 0.5em minus 0.4em\relax Springer, 2016, pp.
  499--515.

\bibitem{deng2019arcface}
J.~Deng, J.~Guo, N.~Xue, and S.~Zafeiriou, ``Arcface: Additive angular margin
  loss for deep face recognition,'' in \emph{Proceedings of the IEEE/CVF
  Conference on Computer Vision and Pattern Recognition}, 2019, pp. 4690--4699.

\bibitem{wang2018cosface}
H.~Wang, Y.~Wang, Z.~Zhou, X.~Ji, D.~Gong, J.~Zhou, Z.~Li, and W.~Liu,
  ``Cosface: Large margin cosine loss for deep face recognition,'' in
  \emph{Proceedings of the IEEE conference on computer vision and pattern
  recognition}, 2018, pp. 5265--5274.

\bibitem{liu2017sphereface}
W.~Liu, Y.~Wen, Z.~Yu, M.~Li, B.~Raj, and L.~Song, ``Sphereface: Deep
  hypersphere embedding for face recognition,'' in \emph{Proceedings of the
  IEEE conference on computer vision and pattern recognition}, 2017, pp.
  212--220.

\bibitem{huang2020curricularface}
Y.~Huang, Y.~Wang, Y.~Tai, X.~Liu, P.~Shen, S.~Li, J.~Li, and F.~Huang,
  ``Curricularface: adaptive curriculum learning loss for deep face
  recognition,'' in \emph{Proceedings of the IEEE/CVF Conference on Computer
  Vision and Pattern Recognition}, 2020, pp. 5901--5910.

\bibitem{liu2019fair}
B.~Liu, W.~Deng, Y.~Zhong, M.~Wang, J.~Hu, X.~Tao, and Y.~Huang, ``Fair loss:
  Margin-aware reinforcement learning for deep face recognition,'' in
  \emph{Proceedings of the IEEE/CVF International Conference on Computer
  Vision}, 2019, pp. 10\,052--10\,061.

\bibitem{huang2008labeled}
G.~B. Huang, M.~Mattar, T.~Berg, and E.~Learned-Miller, ``Labeled faces in the
  wild: A database forstudying face recognition in unconstrained
  environments,'' in \emph{Workshop on faces in'Real-Life'Images: detection,
  alignment, and recognition}, 2008.

\bibitem{sengupta2016frontal}
S.~Sengupta, J.-C. Chen, C.~Castillo, V.~M. Patel, R.~Chellappa, and D.~W.
  Jacobs, ``Frontal to profile face verification in the wild,'' in \emph{2016
  IEEE Winter Conference on Applications of Computer Vision (WACV)}.\hskip 1em
  plus 0.5em minus 0.4em\relax IEEE, 2016, pp. 1--9.

\bibitem{zheng2018cross}
T.~Zheng and W.~Deng, ``Cross-pose lfw: A database for studying cross-pose face
  recognition in unconstrained environments,'' \emph{Beijing University of
  Posts and Telecommunications, Tech. Rep}, vol.~5, p.~7, 2018.

\bibitem{moschoglou2017agedb}
S.~Moschoglou, A.~Papaioannou, C.~Sagonas, J.~Deng, I.~Kotsia, and
  S.~Zafeiriou, ``Agedb: the first manually collected, in-the-wild age
  database,'' in \emph{Proceedings of the IEEE Conference on Computer Vision
  and Pattern Recognition Workshops}, 2017, pp. 51--59.

\bibitem{zheng2017cross}
T.~Zheng, W.~Deng, and J.~Hu, ``Cross-age lfw: A database for studying
  cross-age face recognition in unconstrained environments,'' \emph{arXiv
  preprint arXiv:1708.08197}, 2017.

\bibitem{whitelam2017iarpa}
C.~Whitelam, E.~Taborsky, A.~Blanton, B.~Maze, J.~Adams, T.~Miller, N.~Kalka,
  A.~K. Jain, J.~A. Duncan, K.~Allen \emph{et~al.}, ``Iarpa janus benchmark-b
  face dataset,'' in \emph{proceedings of the IEEE conference on computer
  vision and pattern recognition workshops}, 2017, pp. 90--98.

\bibitem{maze2018iarpa}
B.~Maze, J.~Adams, J.~A. Duncan, N.~Kalka, T.~Miller, C.~Otto, A.~K. Jain,
  W.~T. Niggel, J.~Anderson, J.~Cheney \emph{et~al.}, ``Iarpa janus
  benchmark-c: Face dataset and protocol,'' in \emph{2018 International
  Conference on Biometrics (ICB)}.\hskip 1em plus 0.5em minus 0.4em\relax IEEE,
  2018, pp. 158--165.

\bibitem{zhang2020advkin}
L.~Zhang, Q.~Duan, D.~Zhang, W.~Jia, and X.~Wang, ``Advkin: Adversarial
  convolutional network for kinship verification,'' \emph{IEEE transactions on
  cybernetics}, vol.~51, no.~12, pp. 5883--5896, 2020.

\bibitem{wang2018additive}
F.~Wang, J.~Cheng, W.~Liu, and H.~Liu, ``Additive margin softmax for face
  verification,'' \emph{IEEE Signal Processing Letters}, vol.~25, no.~7, pp.
  926--930, 2018.

\bibitem{hu2014discriminative}
J.~Hu, J.~Lu, and Y.-P. Tan, ``Discriminative deep metric learning for face
  verification in the wild,'' in \emph{Proceedings of the IEEE conference on
  computer vision and pattern recognition}, 2014, pp. 1875--1882.

\bibitem{liu2017adaptive}
X.~Liu, B.~Vijaya~Kumar, J.~You, and P.~Jia, ``Adaptive deep metric learning
  for identity-aware facial expression recognition,'' in \emph{Proceedings of
  the IEEE conference on computer vision and pattern recognition workshops},
  2017, pp. 20--29.

\bibitem{turk1991eigenfaces}
M.~Turk and A.~Pentland, ``Eigenfaces for recognition,'' \emph{Journal of
  cognitive neuroscience}, vol.~3, no.~1, pp. 71--86, 1991.

\bibitem{belhumeur1997eigenfaces}
P.~N. Belhumeur, J.~P. Hespanha, and D.~J. Kriegman, ``Eigenfaces vs.
  fisherfaces: Recognition using class specific linear projection,'' \emph{IEEE
  Transactions on pattern analysis and machine intelligence}, vol.~19, no.~7,
  pp. 711--720, 1997.

\bibitem{ahonen2004face}
T.~Ahonen, A.~Hadid, and M.~Pietik{\"a}inen, ``Face recognition with local
  binary patterns,'' in \emph{Computer Vision-ECCV 2004: 8th European
  Conference on Computer Vision, Prague, Czech Republic, May 11-14, 2004.
  Proceedings, Part I 8}.\hskip 1em plus 0.5em minus 0.4em\relax Springer,
  2004, pp. 469--481.

\bibitem{saeed2024framework}
V.~A. Saeed, ``A framework for recognition of facial expression using hog
  features,'' \emph{International Journal of Mathematics, Statistics, and
  Computer Science}, vol.~2, pp. 1--8, 2024.

\bibitem{kareem2024face}
O.~S. Kareem \emph{et~al.}, ``Face mask detection using haar cascades
  classifier to reduce the risk of coved-19,'' \emph{International Journal of
  Mathematics, Statistics, and Computer Science}, vol.~2, pp. 19--27, 2024.

\bibitem{ngo2020facial}
Q.~T. Ngo and S.~Yoon, ``Facial expression recognition based on
  weighted-cluster loss and deep transfer learning using a highly imbalanced
  dataset,'' \emph{Sensors}, vol.~20, no.~9, p. 2639, 2020.

\bibitem{sun2020circle}
Y.~Sun, C.~Cheng, Y.~Zhang, C.~Zhang, L.~Zheng, Z.~Wang, and Y.~Wei, ``Circle
  loss: A unified perspective of pair similarity optimization,'' in
  \emph{Proceedings of the IEEE/CVF Conference on Computer Vision and Pattern
  Recognition}, 2020, pp. 6398--6407.

\bibitem{liu2016large}
W.~Liu, Y.~Wen, Z.~Yu, and M.~Yang, ``Large-margin softmax loss for
  convolutional neural networks.'' in \emph{ICML}, vol.~2, no.~3, 2016, p.~7.

\bibitem{sohn2016improved}
K.~Sohn, ``Improved deep metric learning with multi-class n-pair loss
  objective,'' in \emph{Advances in neural information processing systems},
  2016, pp. 1857--1865.

\bibitem{li2021adaptively}
H.~Li, N.~Wang, X.~Ding, X.~Yang, and X.~Gao, ``Adaptively learning facial
  expression representation via cf labels and distillation,'' \emph{IEEE
  Transactions on Image Processing}, vol.~30, pp. 2016--2028, 2021.

\bibitem{li2022crs}
H.~Li, N.~Wang, X.~Yang, and X.~Gao, ``Crs-cont: A well-trained general encoder
  for facial expression analysis,'' \emph{IEEE Transactions on Image
  Processing}, vol.~31, pp. 4637--4650, 2022.

\bibitem{li2022towards}
H.~Li, N.~Wang, X.~Yang, X.~Wang, and X.~Gao, ``Towards semi-supervised deep
  facial expression recognition with an adaptive confidence margin,'' in
  \emph{Proceedings of the IEEE/CVF Conference on Computer Vision and Pattern
  Recognition}, 2022, pp. 4166--4175.

\bibitem{hasnat2017mises}
M.~Hasnat, J.~Bohn{\'e}, J.~Milgram, S.~Gentric, L.~Chen \emph{et~al.}, ``von
  mises-fisher mixture model-based deep learning: Application to face
  verification,'' \emph{arXiv preprint arXiv:1706.04264}, 2017.

\bibitem{zhe2019directional}
X.~Zhe, S.~Chen, and H.~Yan, ``Directional statistics-based deep metric
  learning for image classification and retrieval,'' \emph{Pattern
  Recognition}, vol.~93, pp. 113--123, 2019.

\bibitem{li2021spherical}
S.~Li, J.~Xu, X.~Xu, P.~Shen, S.~Li, and B.~Hooi, ``Spherical confidence
  learning for face recognition,'' in \emph{Proceedings of the IEEE/CVF
  Conference on Computer Vision and Pattern Recognition}, 2021, pp.
  15\,629--15\,637.

\bibitem{he2009learning}
H.~He and E.~A. Garcia, ``Learning from imbalanced data,'' \emph{IEEE
  Transactions on knowledge and data engineering}, vol.~21, no.~9, pp.
  1263--1284, 2009.

\bibitem{krawczyk2014cost}
B.~Krawczyk, M.~Wo{\'z}niak, and G.~Schaefer, ``Cost-sensitive decision tree
  ensembles for effective imbalanced classification,'' \emph{Applied Soft
  Computing}, vol.~14, pp. 554--562, 2014.

\bibitem{tang2008svms}
Y.~Tang, Y.-Q. Zhang, N.~V. Chawla, and S.~Krasser, ``Svms modeling for highly
  imbalanced classification,'' \emph{IEEE Transactions on Systems, Man, and
  Cybernetics, Part B (Cybernetics)}, vol.~39, no.~1, pp. 281--288, 2008.

\bibitem{cui2019class}
Y.~Cui, M.~Jia, T.-Y. Lin, Y.~Song, and S.~Belongie, ``Class-balanced loss
  based on effective number of samples,'' in \emph{Proceedings of the IEEE/CVF
  conference on computer vision and pattern recognition}, 2019, pp. 9268--9277.

\bibitem{cao2019learning}
K.~Cao, C.~Wei, A.~Gaidon, N.~Arechiga, and T.~Ma, ``Learning imbalanced
  datasets with label-distribution-aware margin loss,'' \emph{Advances in
  neural information processing systems}, vol.~32, 2019.

\bibitem{mardia2009directional}
K.~V. Mardia and P.~E. Jupp, \emph{Directional statistics}.\hskip 1em plus
  0.5em minus 0.4em\relax John Wiley \& Sons, 2009, vol. 494.

\bibitem{temme1996special}
N.~M. Temme, \emph{Special functions: An introduction to the classical
  functions of mathematical physics}.\hskip 1em plus 0.5em minus 0.4em\relax
  John Wiley \& Sons, 1996.

\bibitem{banerjee2005clustering}
A.~Banerjee, I.~S. Dhillon, J.~Ghosh, S.~Sra, and G.~Ridgeway, ``Clustering on
  the unit hypersphere using von mises-fisher distributions.'' \emph{Journal of
  Machine Learning Research}, vol.~6, no.~9, 2005.

\bibitem{li2019gradient}
B.~Li, Y.~Liu, and X.~Wang, ``Gradient harmonized single-stage detector,'' in
  \emph{Proceedings of the AAAI Conference on Artificial Intelligence},
  vol.~33, no.~01, 2019, pp. 8577--8584.

\bibitem{he2020momentum}
K.~He, H.~Fan, Y.~Wu, S.~Xie, and R.~Girshick, ``Momentum contrast for
  unsupervised visual representation learning,'' in \emph{Proceedings of the
  IEEE/CVF conference on computer vision and pattern recognition}, 2020, pp.
  9729--9738.

\bibitem{grill2020bootstrap}
J.-B. Grill, F.~Strub, F.~Altch{\'e}, C.~Tallec, P.~Richemond, E.~Buchatskaya,
  C.~Doersch, B.~Avila~Pires, Z.~Guo, M.~Gheshlaghi~Azar \emph{et~al.},
  ``Bootstrap your own latent-a new approach to self-supervised learning,''
  \emph{Advances in Neural Information Processing Systems}, vol.~33, pp.
  21\,271--21\,284, 2020.

\bibitem{VGGFace2}
Q.~Cao, L.~Shen, W.~Xie, O.~M. Parkhi, and A.~Zisserman, ``Vggface2: A dataset
  for recognising faces across pose and age,'' in \emph{2018 13th IEEE
  International Conference on Automatic Face Gesture Recognition (FG 2018)},
  2018, pp. 67--74.

\bibitem{deng2018uv}
J.~Deng, S.~Cheng, N.~Xue, Y.~Zhou, and S.~Zafeiriou, ``Uv-gan: Adversarial
  facial uv map completion for pose-invariant face recognition,'' in
  \emph{Proceedings of the IEEE conference on computer vision and pattern
  recognition}, 2018, pp. 7093--7102.

\bibitem{meng2021magface}
Q.~Meng, S.~Zhao, Z.~Huang, and F.~Zhou, ``Magface: A universal representation
  for face recognition and quality assessment,'' in \emph{Proceedings of the
  IEEE/CVF Conference on Computer Vision and Pattern Recognition}, 2021, pp.
  14\,225--14\,234.

\bibitem{xie2018multicolumn}
W.~Xie and A.~Zisserman, ``Multicolumn networks for face recognition,''
  \emph{arXiv preprint arXiv:1807.09192}, 2018.

\bibitem{zhang2019p2sgrad}
X.~Zhang, R.~Zhao, J.~Yan, M.~Gao, Y.~Qiao, X.~Wang, and H.~Li, ``P2sgrad:
  Refined gradients for optimizing deep face models,'' in \emph{Proceedings of
  the IEEE/CVF Conference on Computer Vision and Pattern Recognition}, 2019,
  pp. 9906--9914.

\bibitem{zhang2019adacos}
X.~Zhang, R.~Zhao, Y.~Qiao, X.~Wang, and H.~Li, ``Adacos: Adaptively scaling
  cosine logits for effectively learning deep face representations,'' in
  \emph{Proceedings of the IEEE/CVF Conference on Computer Vision and Pattern
  Recognition}, 2019, pp. 10\,823--10\,832.

\bibitem{ng2014data}
H.-W. Ng and S.~Winkler, ``A data-driven approach to cleaning large face
  datasets,'' in \emph{2014 IEEE international conference on image processing
  (ICIP)}.\hskip 1em plus 0.5em minus 0.4em\relax IEEE, 2014, pp. 343--347.

\bibitem{liu2019adaptiveface}
H.~Liu, X.~Zhu, Z.~Lei, and S.~Z. Li, ``Adaptiveface: Adaptive margin and
  sampling for face recognition,'' in \emph{Proceedings of the IEEE/CVF
  Conference on Computer Vision and Pattern Recognition}, 2019, pp.
  11\,947--11\,956.

\bibitem{wang2020mis}
X.~Wang, S.~Zhang, S.~Wang, T.~Fu, H.~Shi, and T.~Mei, ``Mis-classified vector
  guided softmax loss for face recognition,'' in \emph{Proceedings of the AAAI
  Conference on Artificial Intelligence}, vol.~34, no.~07, 2020, pp.
  12\,241--12\,248.

\bibitem{kemelmacher2016megaface}
I.~Kemelmacher-Shlizerman, S.~M. Seitz, D.~Miller, and E.~Brossard, ``The
  megaface benchmark: 1 million faces for recognition at scale,'' in
  \emph{Proceedings of the IEEE conference on computer vision and pattern
  recognition}, 2016, pp. 4873--4882.

\bibitem{zhang2016joint}
K.~Zhang, Z.~Zhang, Z.~Li, and Y.~Qiao, ``Joint face detection and alignment
  using multitask cascaded convolutional networks,'' \emph{IEEE Signal
  Processing Letters}, vol.~23, no.~10, pp. 1499--1503, 2016.

\bibitem{tai2019towards}
Y.~Tai, Y.~Liang, X.~Liu, L.~Duan, J.~Li, C.~Wang, F.~Huang, and Y.~Chen,
  ``Towards highly accurate and stable face alignment for high-resolution
  videos,'' in \emph{Proceedings of the AAAI Conference on Artificial
  Intelligence}, vol.~33, no.~01, 2019, pp. 8893--8900.

\bibitem{paszke2019pytorch}
A.~Paszke, S.~Gross, F.~Massa, A.~Lerer, J.~Bradbury, G.~Chanan, T.~Killeen,
  Z.~Lin, N.~Gimelshein, L.~Antiga \emph{et~al.}, ``Pytorch: An imperative
  style, high-performance deep learning library,'' \emph{Advances in neural
  information processing systems}, vol.~32, pp. 8026--8037, 2019.

\bibitem{insightface}
I.~Community, ``Insightface,''
  \url{https://github.com/deepinsight/insightface}, 2021, accessed: 2022-01-01.

\bibitem{an2020partical_fc}
X.~An, X.~Zhu, Y.~Xiao, L.~Wu, M.~Zhang, Y.~Gao, B.~Qin, D.~Zhang, and F.~Ying,
  ``Partial fc: Training 10 million identities on a single machine,'' in
  \emph{Arxiv 2010.05222}, 2020.

\bibitem{shi2019probabilistic}
Y.~Shi and A.~K. Jain, ``Probabilistic face embeddings,'' in \emph{Proceedings
  of the IEEE/CVF International Conference on Computer Vision}, 2019, pp.
  6902--6911.

\end{thebibliography}
